\documentclass[conference,preprint]{IEEEtran}
\IEEEoverridecommandlockouts
\usepackage{cite}
\usepackage{bm}
\usepackage{amsmath,amssymb,amsfonts,amsbsy}
\usepackage{textcomp}
\usepackage{xcolor}
\usepackage{soul}
\usepackage{xspace}
\usepackage{mathrsfs}
\usepackage{multirow}
\usepackage{multicol}
\usepackage{enumitem}
\usepackage{dutchcal}
\usepackage{graphicx} 
\usepackage{hyperref}
\usepackage{pifont}
\usepackage{svg}
\usepackage{threeparttable}
\usepackage{color}
\usepackage{booktabs}
\usepackage{diagbox}
\usepackage{algorithm}
\usepackage{algorithmicx}
\usepackage{algpseudocode}
\usepackage{caption}
\usepackage{subcaption}
\usepackage{url}

\def\BibTeX{{\rm B\kern-.05em{\sc i\kern-.025em b}\kern-.08em
    T\kern-.1667em\lower.7ex\hbox{E}\kern-.125emX}}
    
\begin{document}

\newcommand{\argmax}{\operatornamewithlimits{argmax}}
\newcommand{\argmin}{\operatornamewithlimits{argmin}}
\newcommand{\minmaxscale}{\operatornamewithlimits{MinMaxScale}}

\newcommand{\method}{{UADB}\xspace}
\newcommand{\yht}[1]{{\color{blue}{#1}}}
\newcommand{\cw}[1]{{\color{red}{[(CW): #1]}}}
\newcommand{\lzn}[1]{{\color{blue}{[(LZN): #1]}}}

\title{\method: Unsupervised Anomaly Detection Booster\\
\thanks{*The two authors contributed equally to this work. $\dagger$Corresponding authors.}
}

\author{
\IEEEauthorblockA{Hangting Ye\textsuperscript{1*}, Zhining Liu\textsuperscript{2*}, Xinyi Shen\textsuperscript{3}, Wei Cao\textsuperscript{4}, Shun Zheng\textsuperscript{4}, Xiaofan Gui\textsuperscript{4}, Huishuai Zhang\textsuperscript{4}, Yi Chang\textsuperscript{1$\dagger$}, Jiang Bian\textsuperscript{4$\dagger$} \\
\textsuperscript{1}School of Artificial Intelligence, Jilin University, Changchun, China \\
\textsuperscript{2}University of Illinois Urbana-Champaign, Champaign, United States \\
\textsuperscript{3}Renmin University of China, Beijing, China \\
\textsuperscript{4}Microsoft Research, Beijing, China \\
yeht2118@mails.jlu.edu.cn, liu326@illinois.edu, xinyi\_shen@ruc.edu.cn, \\
\{Wei.Cao, shun.zheng, xiaofangui, huishuai.zhang, jiang.bian\}@microsoft.com, yichang@jlu.edu.cn
}
}




\maketitle
\begin{abstract}
\label{sec:abstract}
Unsupervised Anomaly Detection (UAD) is a key data mining problem owing to its wide real-world applications.
Due to the complete absence of supervision signals, UAD methods rely on implicit assumptions about anomalous patterns (e.g., scattered/sparsely/densely clustered) to detect anomalies.
However, real-world data are complex and vary significantly across different domains.
No single assumption can describe such complexity and be valid in all scenarios.
This is also confirmed by recent research that shows no UAD method is omnipotent~\cite{han2022adbench}.
Based on above observations, instead of searching for a magic universal winner assumption, we seek to design a general UAD Booster (UADB) that empowers any UAD models with adaptability to different data.
This is a challenging task given the heterogeneous model structures and assumptions adopted by existing UAD methods.
To achieve this, we dive deep into the UAD problem and find that compared to normal data, anomalies (i) lack clear structure/pattern in feature space, thus (ii) harder to learn by model without a suitable assumption, and finally, leads to (iii) high variance between different learners.
In light of these findings, we propose to (i) distill the knowledge of the source UAD model to an imitation learner (booster) that holds no data assumption, then (ii) exploit the variance between them to perform automatic correction, and thus (iii) improve the booster over the original UAD model.
We use a neural network as the booster for its strong expressive power as a universal approximator and ability to perform flexible post-hoc tuning.
Note that UADB is a model-agnostic framework that can enhance heterogeneous UAD models in a unified way.
Extensive experiments on over 80 tabular datasets demonstrate the effectiveness of UADB.
To facilitate further research, code, figures, and datasets are available at UADB's Github repository\footnote{\url{https://github.com/HangtingYe/UADB}}.
\end{abstract}
\begin{IEEEkeywords}
unsupervised anomaly detection, unsupervised learning, outlier detection
\end{IEEEkeywords}

\section{Introduction}
\label{sec:introduction}
Anomaly detection (AD), also known as Outlier Detection, aims to identify the data objects or behaviors that significantly deviate from the majority \cite{zha2020meta}.
AD is considered as a crucial machine learning problem and has been researched in a variety of fields, including Web and Cyber Security (intrusion detection), Social Network Mining (malicious user/news discovery), and Healthcare (rare disease diagnosis)~\cite{chandola2009anomaly}.
Anomalies in data can be translated into significant actionable information in a wide range of application domains. For example, in computer systems, unusual behaviors may show the presence of malicious activities;
in clinical medicine, abnormal MRI images may indicate the presence of a malignant tumor~\cite{spence2001detection}. 

For its wide applications, AD has been an active research area for several decades \cite{grubbs1969procedures, pang2021deep}, and numerous algorithms have been proposed for AD, including supervised, semi-supervised and unsupervised methods \cite{wolpert1997no,han2022adbench}.
However, ground-truth labels usually need to be manually annotated by domain experts, which is both expensive and time-consuming, and accurately marking all types of abnormal samples is usually unaffordable in practice \cite{chandola2009anomaly}.
Hence, the unsupervised anomaly detection (UAD) methods are the most widely applicable techniques as they do not require any label information \cite{chandola2009anomaly}.
This makes UAD a longtime research hotspot in the field, and new studies continue to appear in recent years \cite{ruff2021unifying}.
In this paper, we focus on \textit{unsupervised anomaly detection on tabular datasets}, which is a very challenging problem and has been the focus of most related works in the literature~\cite{han2022adbench, pang2021deep, pang2018learning, wang2019unsupervised, nguyen2018scalable, zong2018deep, chen2016entity}.

However, despite the extensive research efforts that have been made to address this problem, there still does not exist a single universal winner solution that consistently outperforms other counterparts due to the multifaceted complexity of the task~\cite{han2022adbench}.
Specifically, to achieve accurate UAD on tabular datasets, one faces following fundamental challenges:
\begin{itemize}[leftmargin=0.14in]
    \item \textbf{Unsupervision}: 
    In UAD's problem setting, the label information is completely absent in the training phase.
    Since there are no supervision signals that can provide the model with prior knowledge about the anomalous pattern, UAD models can only detect potential anomalies by making implicit assumptions about the anomaly data instances~\cite{ahmed2016survey}.
    \item \textbf{Assumption Misalignment}:
    Common assumptions adopted by UAD methods include: (i) anomalies occur far from their closest neighbors (neighbor-based, e.g.,\cite{angiulli2002fast,zhang2006detecting}); (ii) anomalies does not belong to any cluster/far away from their closest cluster centroid/belong to small and/or sparse clusters (clustering-based, e.g.,\cite{yu2002findout,smith2002clustering,pires2005using}); (iii) anomalies occur in the low probability regions of a stochastic model (statistical-based, e.g.,\cite{ye2001anomaly,aggarwal2005abnormality,agarwal2007detecting}).
    UAD methods get satisfactory performance when their assumptions hold true, but unfortunately, this is usually not the case in practice.
    For instance, statistical-based UAD assume that the data is generated from a particular distribution, which often does not hold  for high dimensional real datasets.
    The nearest neighbor/clustering-based approach makes specific assumptions about the distribution of anomalies, but they are incompatible and even conflict with each other.
    For above reasons, UAD assumptions can be easily violated in real-world data and result in suboptimal performance~\cite{chandola2009anomaly}.
    \item \textbf{Data Heterogeneity}:
    Furthermore, the heterogeneity, diversity, and complexity of tabular data also pose several great challenges to UAD.
    Unlike image/text/graph data that has natural contexts between pixels/channels/words/nodes, tabular data has no such shared contextual attributes that can help detecting anomalies.
    In tabular datasets, features are not explicitly linked to each other and often show heterogeneity: they may vary significantly in value distribution, range, and even space (e.g., continuous vs. categorical features)~\cite{shenkar2021anomaly}.
    Therefore, even if a UAD method works well in one specific tabular dataset, its underlying assumptions are unlikely to hold in another tabular UAD task, since the two datasets can have very different characteristics.
    This is also confirmed by previous research efforts~\cite{wolpert1997no,han2022adbench} which show that \textit{there is no universal winner for all UAD tasks}.
\end{itemize}

\begin{figure}[t]
\centering
\includegraphics[width=0.9\linewidth]{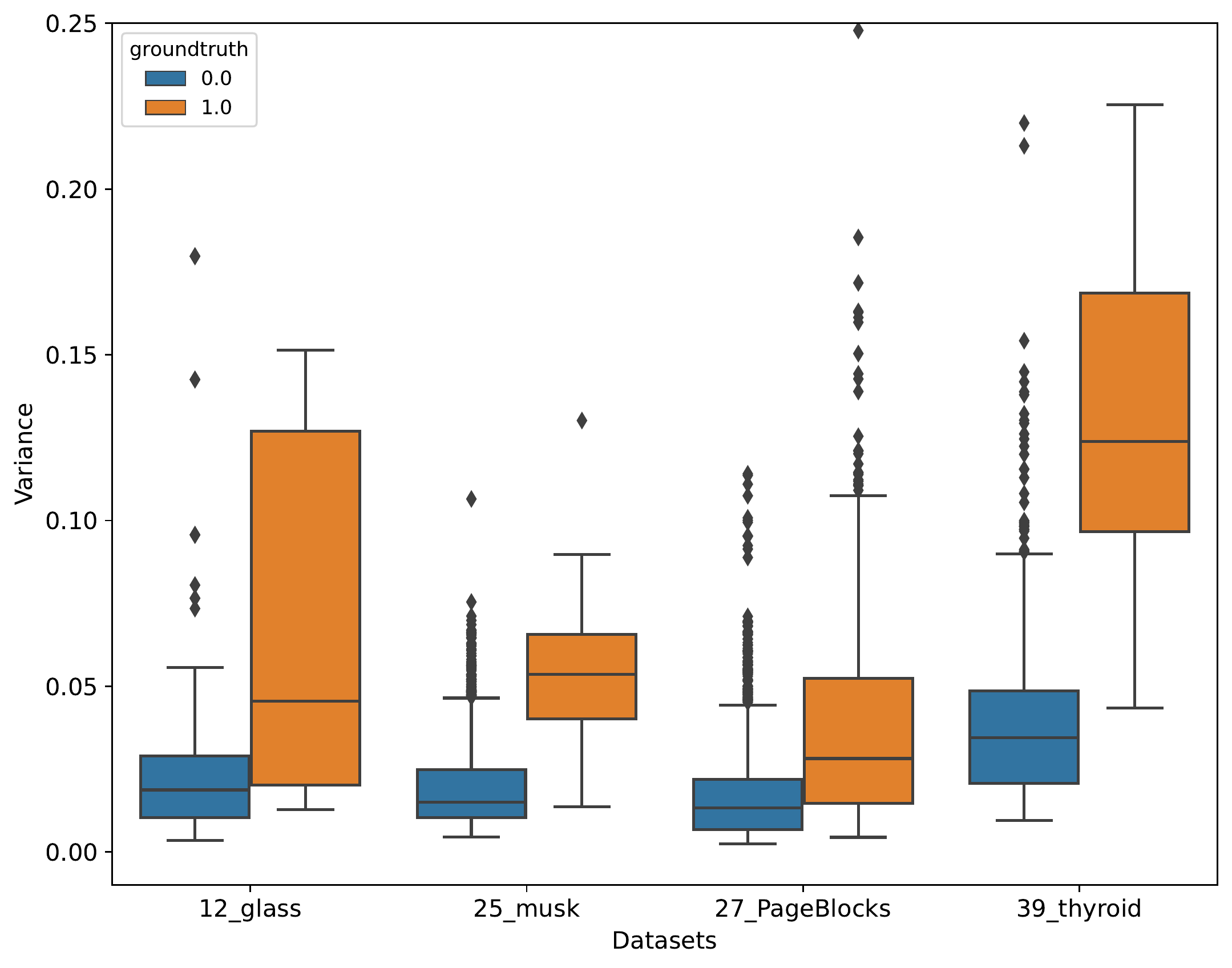}
\caption{
The sample variance of normal (blue) and abnormal (orange) instances in real-world datasets, estimated by an UAD model IForest~\cite{liu2008isolation} and its imitation learner (a MLP trained with IForest's outputs as pseudo labels).
The sample variance is the variance of MLP's prediction and IForest's output, and it is calculated per instance.
Here, groundtruth=1.0 stands for anomalies, while groundtruth=0.0  stands for inliers.
Anomalies consistently demonstrate higher variance across different datasets due to their weak structure/pattern in feature space.
}
\label{fig:std_box}
\vspace{-2em}
\end{figure}

For the above reasons, we believe that the way to better UAD is not to look for a universal winner assumption which is unlikely to exist.
To achieve generally better UAD on diverse and heterogeneous tabular data, the key is to go \textbf{beyond the static assumptions and empower the models with adaptability to different data}.
In this direction, we propose our solution based on two key motivating observations: (i) \textit{the power of proper assumptions} and (ii) \textit{the high variance of anomalies}.
Specifically, (i) \cite{han2022adbench} has shown that, with a proper data assumption, an unsupervised AD method can beat label-informed semi-supervised AD techniques.
This indicates that the data assumption is a powerful tool for detecting specific types of anomalies and therefore should not be discarded outright, but it still needs to be adaptively enhanced so as to find anomalies that do not fit the assumption.
(ii) Besides, unlike classification tasks where each class has a unique underlying distribution, in anomaly detection, anomalies are just irregular instances with no clear structure/pattern in the feature space.
Therefore, compared to normal samples, anomalies are harder to learn by a model and are likely to induce a high variance, i.e., different models' predictions of anomalies will vary, see an example in Fig.~\ref{fig:std_box}.
Such property can be used to support the adaptive augmentation of UAD models.

In light of these analysis, we propose \method (\underline{U}nsupervised \underline{A}nomaly \underline{D}etection \underline{B}ooster), a surprisingly simple yet effective framework that can \textit{boost the prediction accuracy of any mainstream UAD methods on all types of tabular datasets}.
Specifically, given any UAD model $f_T(\cdot)$,  we introduce a booster model $f_B(\cdot)$, then iteratively (i) train $f_B$ in a supervised manner using $f_T$'s predictions as pseudo labels $\hat{\textbf{y}}$; (ii) estimate sample variances $\textbf{v}$ using the output of $f_T$ and $f_B$; (iii) perform error correction by exploiting the sample variance.
Note that having the observation in Fig.~\ref{fig:std_box}, the error correction can be simply done by adding sample variance to the pseudo labels. 
This will result in a large increase in the anomaly score (higher score indicates higher confidence that the sample is an anomaly) for false negatives (mispredicting anomalies) but a small increment for false positives (mispredicting normal samples), thus gradually narrowing the prediction gap between them, and finally altering their rankings in the pseudo labels to achieve error correction.
We thus obtain a booster model $f_B$ that is improved over its prototype $f_T$ as it benefits from both knowledge distillation and adaptive error correction.
We highlight that \method does not make any assumptions about the input UAD prototype $f_T$, it is a model-agnostic framework that can generally enhance any $f_T$ in a unified way.
Extensive experiments and analyses conducted on over 80 tabular UAD datasets demonstrate the effectiveness of UADB.

To sum up, this paper makes following 3-fold contributions:
\begin{itemize}[leftmargin=*]
    \item 
    We investigate the key challenges of Unsupervised Anomaly Detection (UAD) on tabular data, such as assumption misalignment and data heterogeneity. They prompt us to explore a new direction: empower static-assumption-based UAD models with adaptability to different data.
    \item
    We propose UADB, a model-agnostic framework that can effectively boost any UAD model's performance on tabular datasets via knowledge distillation and adaptive error correction. To our best knowledge, UADB is the first of its kind as a general augmentation framework for UAD models.
    \item
    We conducted extensive experiments on more than 80 tabular datasets along with comprehensive analysis and visualization. These results validate the effectiveness of the proposed UADB framework and may provide valuable insights for further research on versatile UAD boosters.
\end{itemize}

The rest of this paper is organized as follows:
Section~\ref{sec:related work} reviews closely related works in unsupervised anomaly detection and knowledge distillation. 
Section~\ref{sec:methodology} introduces the notations and describes the proposed \method framework.
Section~\ref{sec:experiment} presents the experimental results as well as related discussions and analysis.
And finally, section~\ref{sec:conclusion} concludes the paper.

\section{Related Work}
\label{sec:related work}
In this section, we provide a systematical review of the existing works related to unsupervised anomaly detection and knowledge distillation applications.

\subsection{Unsupervised Anomaly Detection}
Anomaly detection is a big topic in machine learning, including supervised, semi-supervised, and unsupervised methods.
Due to the advantage of not requiring ground-truth labels, unsupervised anomaly detection methods are widely used.
In the setting of unsupervised anomaly detection, we have no prior knowledge about which type of data is normal or which is abnormal, i.e. the training data is without true labels.
Our task is to find instances that deviate the most from the other instances among all dataset \cite{guthrie2007unsupervised}.
Since unsupervised anomaly detection has drawn interest in the academic community \cite{eskin2002geometric, leung2005unsupervised}, numerous unsupervised anomaly detection methods have been proposed.
These methods could be roughly grouped into shallow and deep methods, with the details as follows.

We list some representative shallow methods: (i) Isolation Forest (IForest) \cite{liu2008isolation} builds an ensemble of trees for a given data set, then use the distance of instance to the root as anomaly score; 
(ii) Histogram-based Outlier Score (HBOS) \cite{goldstein2012histogram}. 
The basic assumption is that the dataset's dimensions are independent. 
Each dimension would be divided into intervals. 
The higher density represents the lower anomaly score;
(iii) Empirical-Cumulative-distribution-based Outlier Detection (ECOD) \cite{li2022ecod} first computes the empirical cumulative distribution for each dimension of the input data.
Then for each dimension, ECOD aggregates the tail probabilities to compute the anomaly score. 

Some representative deep methods are as follows: (i) Deep Support Vector Data Description (DeepSVDD) \cite{ruff2018deep} trains a neural network while minimizing the volume of a hypersphere that encloses the network representations of the data, and the distance of the transformed embedding to the hypersphere's center is used to calculate the anomaly score;
(ii) Deep Autoencoding Gaussian Mixture Model (DAGMM) \cite{zong2018deep} jointly optimizes the parameters of the deep autoencoder and the mixture model simultaneously in an end-to-end fashion, leveraging a separate estimation network to help with the parameter learning of the mixture model.  
The joint optimization eliminates the need for pre-training by assisting the autoencoder escape from less attractive local optima and further reducing reconstruction errors.

\subsection{Knowledge Distillation}
Knowledge Distillation (KD) is a family of techniques that aim to transfer knowledge from a trained source (teacher) model(s) to a target (student) model \cite{hinton2015distilling}.
The student models are usually smaller but perform similarly or even better than the large teacher models.
Such a training scheme is also known as the teacher-student architecture and has been proven to be effective in numerous applications \cite{phuong2019towards, cho2019efficacy}. 
Starting from the success in image classification~\cite{li2017learning,wang2019deepvid,zhu2019low} and other visual recognition tasks~\cite{kong2019cross,yan2019vargfacenet}, more knowledge distillation systems are designed for broader applications such as neural machine translation~\cite{hahn2019self,zhou2019understanding}, feature selection~\cite{fan2020autofs, fan2021interactive}, text generation~\cite{chen2019distilling}, speech recognition~\cite{oord2018parallel,kwon2020adaptive,shen2020knowledge} and so on.

A number of research efforts have explored the application of knowledge distillation to anomaly detection tasks.
\textit{Salehi et al.}\cite{salehi2021multiresolution} tried to explore the multi-layer feature information in distillation, so as to better exploit teacher model's multi-resolution knowledge and get a better student network for image anomaly detection task.
Also for image anomaly detection, \textit{Bergmann et al.} \cite{bergmann2020uninformed} proposed to include multiple student networks and learn from the teacher model's knowledge in an ensemble mannner.
\textit{Wang et al.} \cite{wang2021student} designed a multi-scale feature matching strategy to enable student learning with hierarchical supervision, thus improve the detection accuracy of anomalies of various sizes in images.
Motivated by the fact that practitioners often build a large number of UAD models rather than a single model for reliable further combination and analysis, \textit{Zhao et al.} \cite{zhao2021suod} developed a system for accelerating UAD with large-scale heterogeneous models.
They train a pseudo-supervised simple regressor student model to approximate the large ensemble of heterogeneous UAD models, thus accelerating the inference.

We note that there are several fundamental differences between our proposed UADB framework and the aforementioned KD + UAD techniques:
Many methods are specifically designed for distilling knowledge from large multi-layere neural networks (e.g.,~\cite{salehi2021multiresolution,wang2021student}), and for a specific AD task, such as detecting pixel-level anomaly objects in images.
While in this work, we consider a more general case, to improve any teacher UAD model on heterogeneous tabular datasets.
Furthermore, existing works directly transfer the knowledge from the teacher model without modification and let the student mimic the teacher's behaviors.
In UADB, we estimate the variance using the discrepancy between teacher and student and exploit this information to adjust the pseudo labels, thus achieving adaptive error correction in knowledge distillation.
The details of UADB will be covered in the next section.

\section{Methodology}
\label{sec:methodology}
\begin{figure*}[t]
\centering
\includegraphics[width=\linewidth]{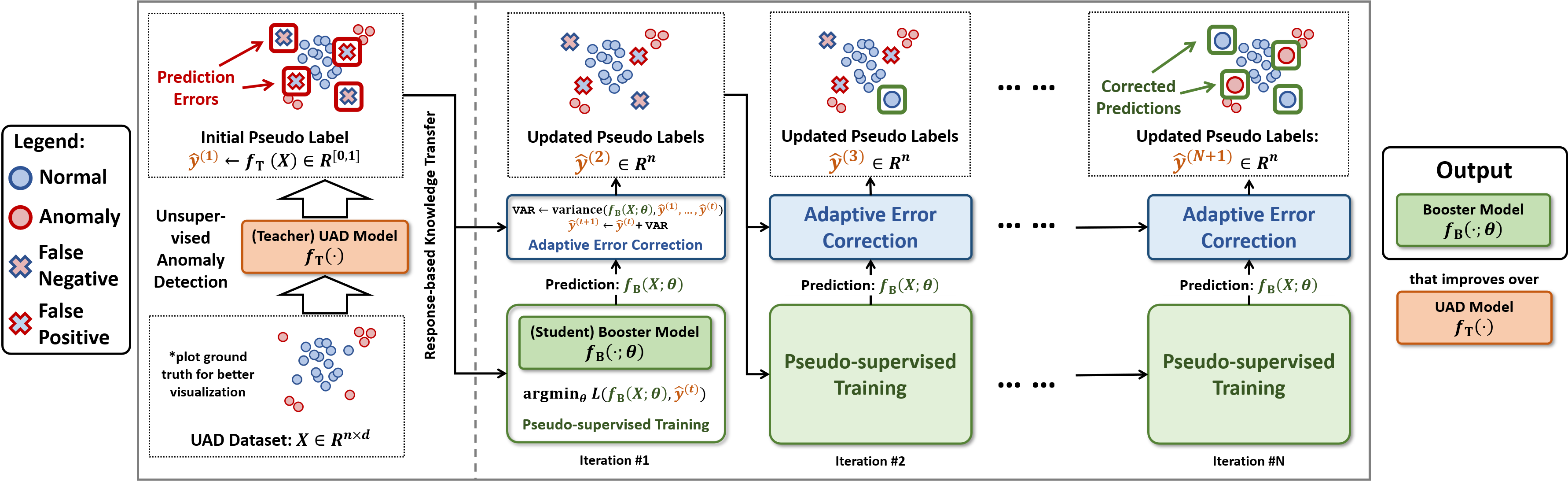}
\caption{Overview of the proposed \method framework. Best viewed in color.}
\label{fig:framework}
\vspace{-1em}
\end{figure*}

In this section, we first introduce the notations and formalize the unsupervised anomaly detection booster problem considered in this paper.
We then demonstrate why anomalies are likely to have high variance and show empirical evidence collected from 80 tabular datasets.
After that, we present how UADB exploits information from both the teacher model and the variance to achieve effective knowledge transfer and error correction.
Finally, we formalize UADB in Algorithm~\ref{algorithm}, 
and Fig~\ref{fig:framework} gives an overview of the UADB framework.

\subsection{Notations and problem definition}
\label{sec:preliminaries}

\begin{table}[h]
\caption{Notation Definitions}
\centering
\begin{tabular}{c|c} 
\toprule
\textbf{Notation} & \textbf{Definition}         \\ 
\midrule
$d$                 & Number of data features  \\ 
\hline
$\textbf{x}:[x_1, x_2,\cdots, x_d]$                 & A data sample  \\ 
\hline
$n$                 & Number of data samples  \\ 
\hline
$\textbf{X} \in \mathbb{R}^{n\times d}$                 & A UAD dataset  \\ 
\hline
$f_S(\cdot): \textbf{x} \to \mathbb{R}^{[0,1]}$        & Source UAD (teacher) model \\ 
\hline
$f_B(\cdot;\theta): \textbf{x} \to \mathbb{R}^{[0,1]}$ & Target Booster (student) model \\
\hline
$\theta$        & Parameters of the booster model \\
\hline
$T$             & Number of student training steps \\
\hline
$\hat{y}_i$        & Pseudo label of the $i$-th sample \\
\hline
$\hat{\textbf{y}}: [\hat{y}_1, \hat{y}_2,\cdots, \hat{y}_n]\in \mathbb{R}^{n}$ &  Pseudo label vector\\
\hline
$\hat{\textbf{y}}^{(t)}: [\hat{y}^{(t)}_1, \hat{y}^{(t)}_2,\cdots, \hat{y}^{(t)}_n]\in \mathbb{R}^{n}$ & Pseudo label vector at $t$-th iteration\\
\bottomrule
\end{tabular}
\end{table}

\textbf{Notations.}
We first introduce the  notations used in this paper.
Let $d$ be the number of input features, a data sample can be represented by its feature vector $\textbf{x}:[x_1, x_2,\cdots, x_d]$, then we can denote a dataset for UAD as $\textbf{X} \in \mathbb{R}^{n\times d}$, where $n$ is the number of samples in the dataset.
Note that there is no ground truth label $y$ in the unsupervised setting.
The goal of unsupervised anomaly detection is to learn a detection model $f(\cdot)$ without ground truth labels. 
A model $f(\cdot)$ takes a feature vector $\textbf{x}$ as input and outputs $\hat{y}\in \mathbb{R}^{[0,1]}$, i.e., the predicted anomaly score of $\textbf{x}$, higher score indicates higher confidence that $\textbf{x}$ is an anomaly.

\textbf{Problem Definition.}
In this paper, we consider the problem of finding a booster model $f_B(\cdot)$ for a given source UAD model $f_S(\cdot)$.
As described before, this is achieved by iterative knowledge distillation with error correction by estimating and exploiting the variance between teacher and student models.
Formally, given a source UAD model $f_S(\cdot): \textbf{x} \to \mathbb{R}^{[0,1]}$, we consider a parameterized booster model $f_B(\cdot;\theta): \textbf{x} \to \mathbb{R}^{[0,1]}$ with $\theta$ denoting its parameters.
The goal of learning a UAD booster model is to find a parameter set $\theta^*$ that maximize $f_B(\cdot;\theta)$'s prediction accuracy on $\textbf{X}$ given the source UAD model $f_S(\cdot)$ and dataset $\textbf{X}$.

More specifically, we use the predictions of the source model $f_S(\textbf{X})$ as the initial pseudo label vector $\hat{\textbf{y}}$ for knowledge distillation.
The target of distillation, booster model $f_B(\cdot;\theta)$ is a neural network.
Note that unlike typical knowledge distillation settings, the source knowledge, i.e., $\hat{\textbf{y}}$ will be adjusted for error correction during the booster model $f_B(\cdot;\theta)$'s pseudo-supervised training process.
Suppose the total number of training steps is $T$, we denote the adjusted pseudo label vector at step/iteration $t$ as $\hat{\textbf{y}}^{(t)}$.
Let $\mathcal{L}$ be the loss function, then in each training step, we need to (i) optimize $\theta$ to minimize $\mathcal{L}(f_B(\textbf{X};\theta), \hat{\textbf{y}}^{(t)})$ for knowledge transfer, and (ii) update $\hat{\textbf{y}}^{(t)}$ for error correction.
The former is a standard supervised learning objective that can be handled by many optimizers, but the latter objective has no straightforward solution.
We will introduce our solution in the following sections.

\subsection{Anomalies pattern on variance}
\textbf{Motivations.}
In order to perform error correction, we need information or statistics that can distinguish between abnormal and normal samples.
Normally such information can be obtained from (partial) ground truth labels or prior knowledge about the anomalies' pattern provided by domain experts. But unfortunately, none of them is available in UAD.
We have to find a new discriminative feature for error correction.

To achieve this, we look back to the fundamental difference between anomaly detection tasks and classification tasks. In classification, each class has a unique underlying distribution/structure in the feature space, and classifiers can distinguish them by learning the difference between class distributions. While in anomaly detection, only normal data has a meaningful underlying structure, and anomalies are just "abnormal" instances without a clear pattern, as they can be caused by multiple unknown, hidden, and even random factors (e.g., data corruption, sensor failure)~\cite{ahmed2016survey}.
This is also the reason why one-class learning (only learning from the pattern of normal data) obtains great success in anomaly detection.

Now we know that compared to normal samples, anomalies lack a clear structure/pattern in the feature space, which can be a sharp knife for distinguishing anomalies.
Specifically, data samples that lack a clear pattern usually have high variance in predictions since they are hard to fit by a simple hypothesis (model).
Therefore, anomalies should likely have high variances, i.e., different models' predictions for one anomaly may vary significantly, and this property can be used as a discriminative feature to support error correction.

\begin{figure}[h]
\centering
\includegraphics[width=0.9\linewidth]{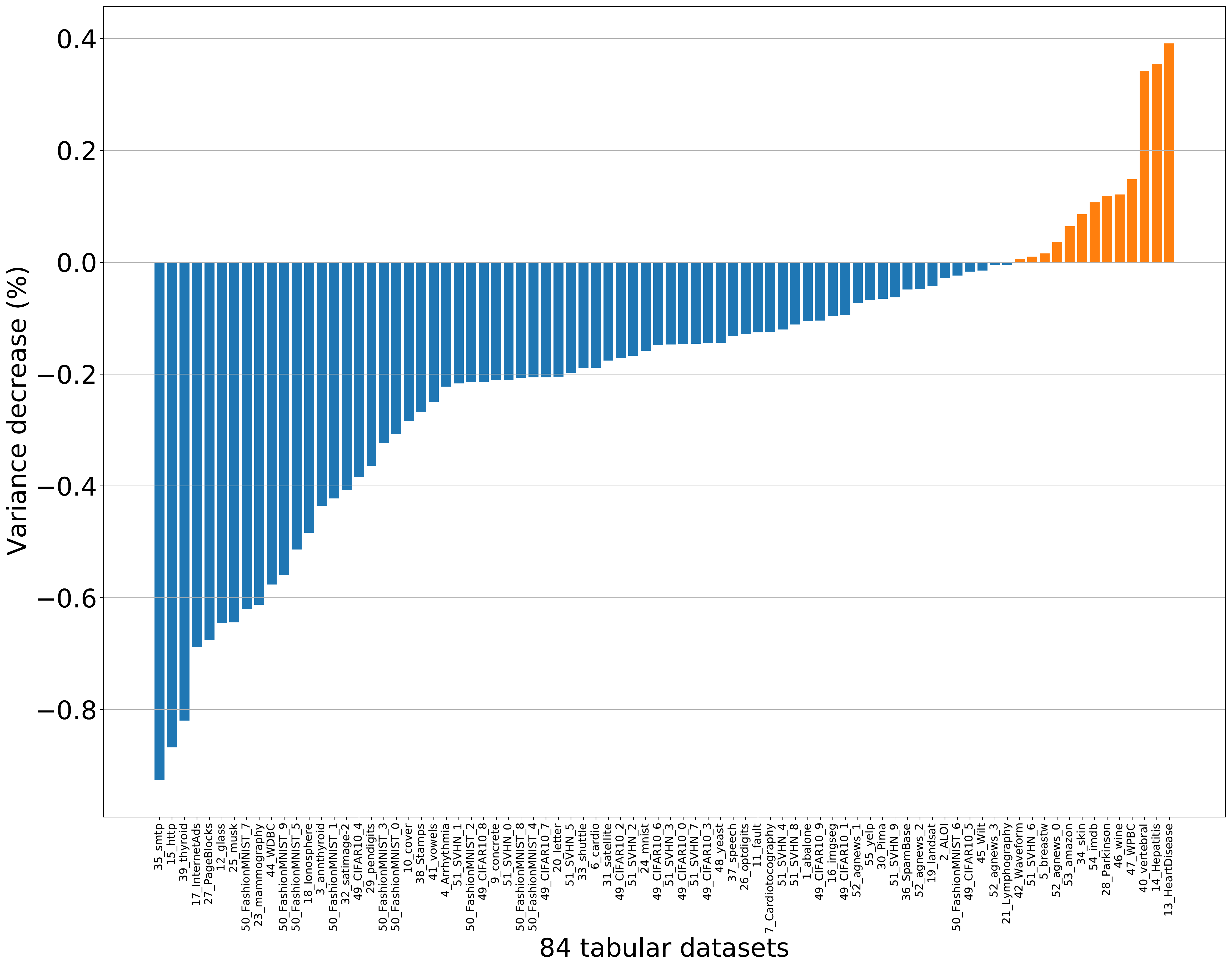}
\caption{
Comparison of average variance of normal samples and anomalies on 84 different tabular datasets.
Negative value indicates that anomalies have higher average variance compared to normal samples, which holds true on 71 out of 84 datasets.
}
\label{fig:std_statistic}
\end{figure}

\textbf{Empirical Evidence.}
To verify whether the variance can be used for error correction, we examine 84 real-world tabular datasets (described in Table~\ref{tab:datasets}) to collect some empirical evidence. 
Specifically, for each dataset, we train a teacher UAD model $f_S$ on it, then use its predictions as static labels to train a pseudo-supervised student model $f_B$.
We estimate the sample variance by measuring the difference between the predictions of the teacher and the student model $\hat{v}_i=variance([f_S(\textbf{x}_i), f_B(\textbf{x}_i)])$ ($\textbf{x}_i$ is the $i$-th sample).
We then compute the average variance of normal samples and anomalies, respectively, i.e., $\hat{v}_\text{normal/abnormal}=\sum_{i\in \text{normal/abnormal}}(\hat{v}_i)/S$, $S$ is a normalization term that equals to the number of normal/abnormal instances.
To improve the visual display of the results, we show the relative average variance difference between normal and anomalies, i.e., $(\hat{v}_\text{normal}-\hat{v}_\text{abnormal})/\hat{v}_\text{abnormal}$.
Negative value indicates that anomalies have higher average variance than normal samples.
Results are shown in Fig.~\ref{fig:std_statistic}.

As observed in Fig.~\ref{fig:std_statistic}, compared to normal samples, anomalies have higher average variance on 85\% (71/84) tabular datasets, which directly validates our previous thoughts.
The relative differences are considerably high ($>5\%$) on 60/84 datasets.
We note that the \textit{variance can be naturally estimated in the teacher-student architecture}, between teacher and student model and/or student model checkpoints at different steps.
This allows us to exploit the variance difference between normal samples and anomalies, achieving dynamic error correction during knowledge distillation.


\subsection{Knowledge transfer and error correction}
\label{sec:correct mechanism}
\textbf{UADB Procedure.}
With the above analysis, we now elaborate on the technical details of UADB training procedure.
As discussed before, appropriate data assumptions are powerful tools for detecting specific types of anomalies and therefore should not be discarded outright, but they need to be enhanced with adaptive error correction so as to handle anomalies that do not fit the assumption.
Accordingly, UADB is designed to:
\begin{itemize}
    \item keep the prior knowledge of the UAD model and its assumption by knowledge transfer;
    \item perform adaptive error correction during transfer by exploiting the sample variance at the same time.
\end{itemize}

Specifically, given a source UAD model $f_S(\cdot): \textbf{x} \to \mathbb{R}^{[0,1]}$ fitted on dataset $\textbf{X}$, we consider a parameterized booster model $f_B(\cdot;\theta): \textbf{x} \to \mathbb{R}^{[0,1]}$ with $\theta$ denoting its parameters.
In UADB, we use a neural network as the parameterized booster model for its strong expressive power as a universal approximator, this is important for handling diverse UAD models with different architectures.
We use the source model's predictions as the initial pseudo label vector $\hat{\textbf{y}}^{(1)}$, i.e., $\hat{\textbf{y}}^{(1)} = f_S(\textbf{X})$.
Then in $t$-th iteration, repeat:
\begin{itemize}
    \item Update $\theta$ w.r.t objective $\argmin_\theta\mathcal{L}(f_B(\textbf{X};\theta), \hat{\textbf{y}}^{(t)})$;
    \item Estimate variance vector $\hat{\textbf{v}}$ using $f_B(\textbf{X};\theta)$ and all previous pseudo label vectors $\widehat{\textbf{Y}}:\{\hat{\textbf{y}}^{(1)}, \cdots, \hat{\textbf{y}}^{(t)}\}$ (calculated per instance);
    \item Update the pseudo label simply by:
    \begin{itemize}
        \item Adding variance $\hat{\textbf{v}}$ to the pseudo label $\hat{\textbf{y}}^{(t)}$
        \item Renormalize to guarantee $\hat{y_i}\in \mathbb{R}^{[0,1]}, \forall i$
        \item i.e., $\hat{\textbf{y}}^{(t+1)}=\minmaxscale(\hat{\textbf{y}}^{(t)}+\hat{\textbf{v}})$
    \end{itemize}
\end{itemize}
Note that the error correction mechanism in UADB is surprisingly simple.
More complex error correction mechanisms can be designed in many different ways, but following Occam's razor principle of parsimony, we prefer a simpler solution for its elegance, interpretability, and wide applicability.
Algorithm~\ref{algorithm} formalizes the proposed UADB framework.
\begin{algorithm}
\caption{Unsupervised Anomaly Detection Booster}
\label{algorithm}
\begin{algorithmic}[1]
\Require UAD dataset $\textbf{X}$, source UAD model $f_S(\cdot):\textbf{x} \to [0,1]$, number of booster training steps $T$
\State \textbf{Initialize:} 
\State target booster model $f_B(\cdot;\theta)$
\State pseudo label vector $\hat{\textbf{y}}^{(1)}\gets f_S(\textbf{X})$
\State pseudo label matrix $\widehat{\textbf{Y}}\gets [\hat{\textbf{y}}^{(1)}]\in \mathbb{R}^{n\times1}$
\For{$t \gets 1$ to $T$}
\State train $f_B(\cdot;\theta)$ by $\argmin_\theta\mathcal{L}(f_B(\textbf{X};\theta), \hat{\textbf{y}}^{(t)})$
\State compute variance $\hat{\textbf{v}}\gets variance([\widehat{\textbf{Y}}, f_B(\textbf{X};\theta)])$ \yht{\footnotemark}
\State update $\hat{\textbf{y}}^{(t+1)}\gets \minmaxscale(\hat{\textbf{y}}^{(t)}+\hat{\textbf{v}})$
\State update $\widehat{\textbf{Y}}\gets[\widehat{\textbf{Y}},\hat{\textbf{y}}^{(t+1)}]\in\mathbb{R}^{n\times (t+1)}$
\EndFor
\State \Return the booster UAD model $f_B(\cdot;\theta)$
\end{algorithmic}
\end{algorithm}
\footnotetext{Calculated per instance.}

\textbf{Variance for Error Correction.}
To better illustrate why this simple pseudo-label updating rule can achieve error correction, we now present related discussions and case studies.

As demonstrated before, compared to normal data instances, anomalies are likely to have higher prediction variance.
Since UADB adopts a teacher-student architecture for booster model training, the variance can be naturally estimated using the source model $f_S$ and the target model $f_B$.
However, we do not simply compute the variance using only the 2 entries from $f_S$ and $f_B$, since the student model is being updated during the training process, which makes the variance estimation vulnerable to unknown training dynamics.
Inspired by the concept of self-teaching in knowledge distillation research~\cite{gou2021knowledge}, we record all the pseudo-label vectors ($\hat{\textbf{y}}$) in previous steps and compute the variance on all $\hat{\textbf{y}}$ and the current student prediction $f_B(\textbf{X};\theta)$.
This provides us with a more reliable variance estimation by including more predictions from model checkpoints, while mitigating the influence of random training dynamics by using previous $\hat{\textbf{y}}$s instead of student model outputs.
But even with reliable variance estimation $\hat{\textbf{v}}$, why UADB can correct errors by simply updating $\hat{\textbf{y}}^{(t+1)}\gets \minmaxscale(\hat{\textbf{y}}^{(t)}+\hat{\textbf{v}})$?

\textbf{Case Study.}
Let us consider four cases for the source UAD model $f_T(\cdot)$ (without loss of generality, we assume anomalies are positive, i.e., $y=1$ in the ground truth label and the detection threshold is $0.5$):
(i) True Positive (TP): $y=1, \hat{y}_{f_T}=1$; (ii) False Negative (FN): $y=1, \hat{y}_{f_T}=0$; (iii) False Positive (FP): $y=0, \hat{y}_{f_T}=1$; (iv) True Negative (TN): $y=0, \hat{y}_{f_T}=0$.
Table~\ref{tab:4 cases} summarizes the four cases.

\begin{table}[h]
\centering
\caption{Four types of instances in UAD.}
\label{tab:4 cases}
\begin{tabular}{ccccc} 
\toprule
\textbf{Case} & \textbf{Ground Truth} & \textbf{Label} & \textbf{Prediction} & \textbf{Variance} \\ \midrule
TP & abnormal & $y=1$ & $\hat{y}_{f_T}=1$ & high \\
FN & abnormal & $y=1$ & $\hat{y}_{f_T}=0$ & high \\
FP & normal & $y=0$ & $\hat{y}_{f_T}=1$ & low \\
TN & normal & $y=0$ & $\hat{y}_{f_T}=0$ & low \\
\bottomrule
\end{tabular}
\end{table}

It is straightforward to observe from the Table~\ref{tab:4 cases} that the goal of error correction is to correct FP and FN in the pseudo labels.
Specifically, in the booster model $f_B(\cdot)$, we want to increase its prediction of FN while decreasing its prediction of FP.
Also, note that the variance of FN is more likely to be higher than FP, so intuitively, adding the respective variance to the pseudo-label will naturally reduce the error gap between FN and FP.
This procedure can be repeated until the gap between FN and FP is eliminated and their relative relationship is inverted in the pseudo-label values, i.e., errors are corrected.
We now discuss these cases in detail. 

\subsubsection{True Positive (TP) and True Negative (TN)}
For a TP instance $\textbf{x}_\text{TP}$, the initial pseudo label $\hat{y}_\text{TP}=f_T(\textbf{x}_\text{TP})$ is close to 1, and it has a high variance $\hat{v}_\text{abnormal}$.
After adding the variance, $\hat{y}_\text{TP} (\to 1) + \hat{v}_\text{abnormal} (>\hat{v}_\text{normal})$ is likely to be a large value that greater than any other types of instances (FN, FP, TN), so after min-max scaling, its new pseudo label $\hat{y}^*_\text{TP}$ will still be close to 1.
Oppositely, for a TN instance $\textbf{x}_\text{TN}$, the initial pseudo label $\hat{y}_\text{TN}=f_T(\textbf{x}_\text{TN})$ is close to 0, and it has a low variance $\hat{v}_\text{normal} < \hat{v}_\text{abnormal}$.
Therefore, after adding its variance, $\hat{y}_\text{TN} (\to 0) + \hat{v}_\text{normal} (<\hat{v}_\text{abnormal})$ will still be the smallest values among all cases.
After min-max scaling, its new pseudo label $\hat{y}^*_\text{TN}$ will still be close to 0.
Thus the correct knowledge in $f_T(\cdot)$ will be maintained.

\subsubsection{False Positive (FP) and False Negative (FN)}
The key function of error correction is to correct the pseudo labels of FPs and FNs.
Specifically, for a FP instance $\textbf{x}_\text{FP}$, its ground truth label $y_\text{FP}=0$ but the pseudo label $\hat{y}_\text{FP}=f_T(\textbf{x}_\text{FP})$ is close to 1.
Thus we want to gradually decrease its score for error correction and expect a smaller score after updating, i.e., $\hat{y}^*_\text{FP} < \hat{y}_\text{FP}$.
To prove this, let's jointly consider FP and TP.
After adding variance, we have the unnormalized scores, 
which is $s_\text{FP}:\hat{y}_\text{FP} (\to 1) + \hat{v}_\text{normal} (<\hat{v}_\text{abnormal})$ for FP and $s_\text{TP}:\hat{y}_\text{TP} (\to 1) + \hat{v}_\text{abnormal} (>\hat{v}_\text{normal})$ for TP.
Note that although the pseudo labels of FP and TP are both close to 1, after adding the variance term, FP's unnormalized score will be smaller than TP due to its low variance, i.e., $s_\text{FP}<s_\text{TP}$.
Then after the scaling, we have $\hat{y}^*_\text{FP}=\frac{s_\text{FP}-s_\text{TN}}{s_\text{TP}-s_\text{TN}}<\hat{y}_\text{FP} (\hat{y}_\text{FP} \to 1)$, which means $\hat{y}_\text{FP}$ will decrease in updates.
Likewise, a FN instance's updated unnormalized score will be larger than that of TN due to its larger variance term, i.e., $s_\text{FN}>s_\text{TN}$, thus $\hat{y}^*_\text{FN}=\frac{s_\text{FN}-s_\text{TN}}{s_\text{TP}-s_\text{TN}}>\hat{y}_\text{FN} (\hat{y}_\text{FN} \to 0)$.
Hence, the error gap between FN and FP can be narrowed by repeatedly applying this updating rule, thus finally achieving error correction.

\begin{figure}[t]
\centering
\begin{subfigure}[b]{0.475\linewidth}
\centering
\includegraphics[width=\textwidth]{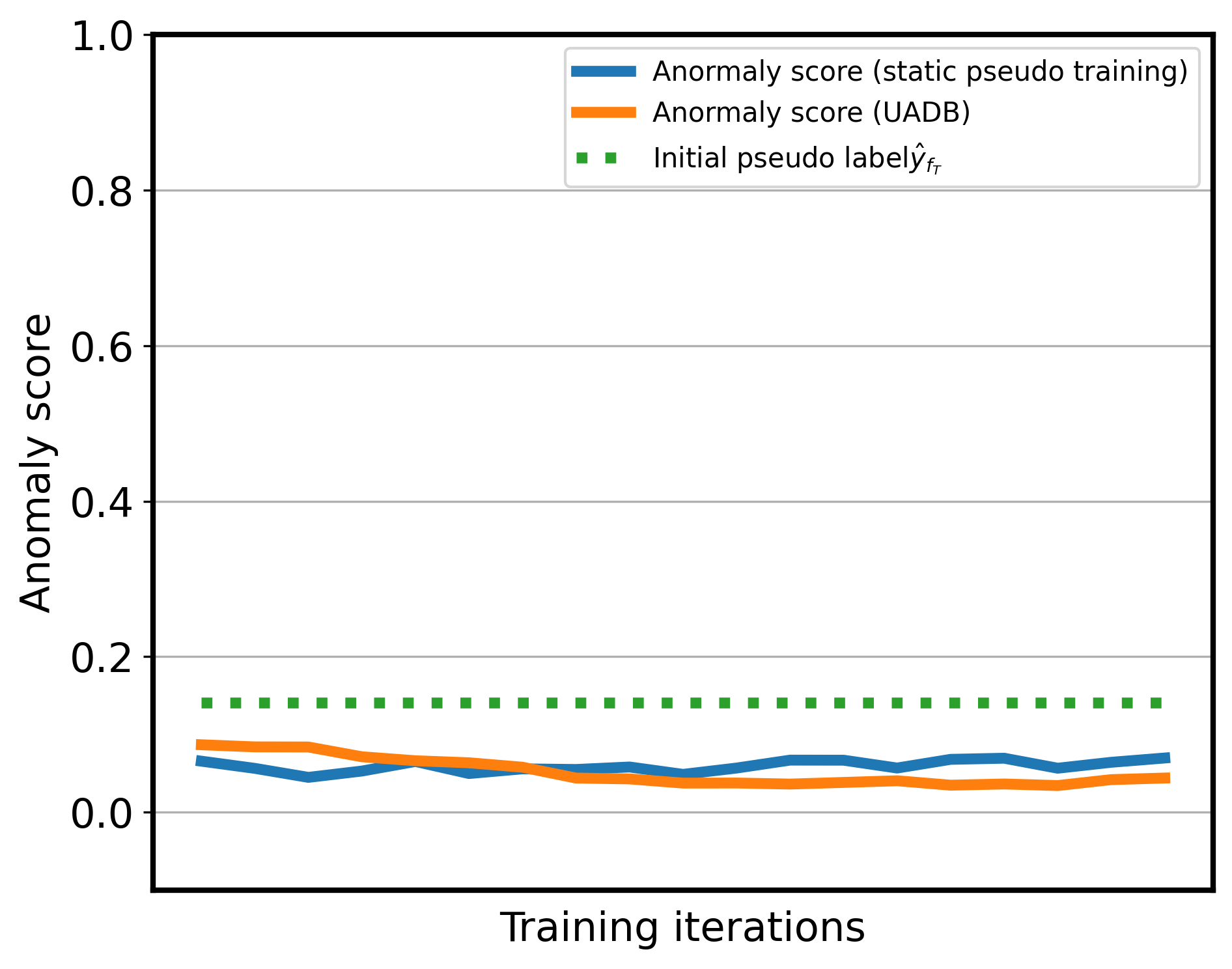}
\caption{True Negative (TN)}
\label{fig:00}
\end{subfigure}
\begin{subfigure}[b]{0.475\linewidth}
\centering
\includegraphics[width=\textwidth]{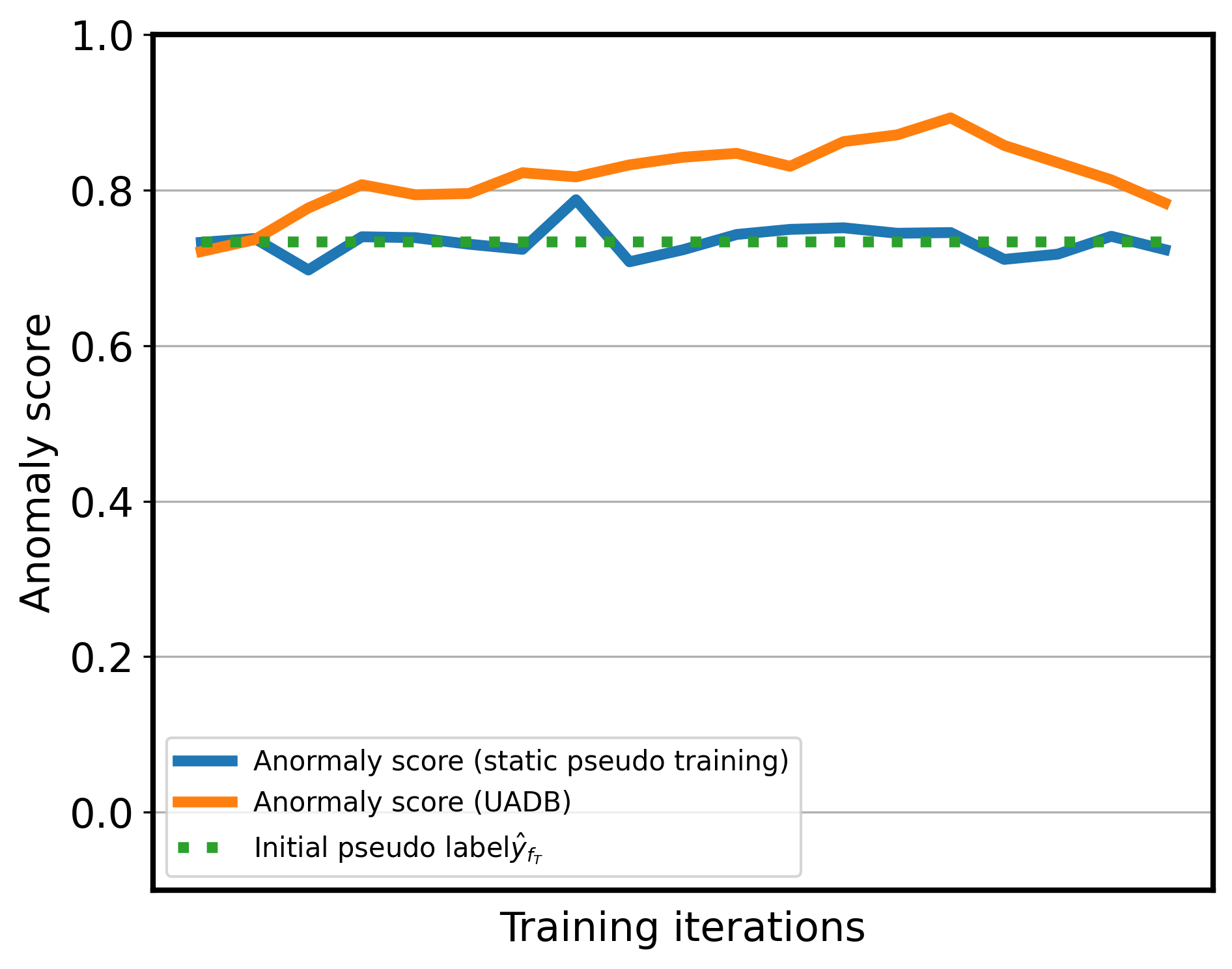}
\caption{True Positive (TP)}
\label{fig:11}
\end{subfigure}
\begin{subfigure}[b]{0.475\linewidth}
\centering
\includegraphics[width=\textwidth]{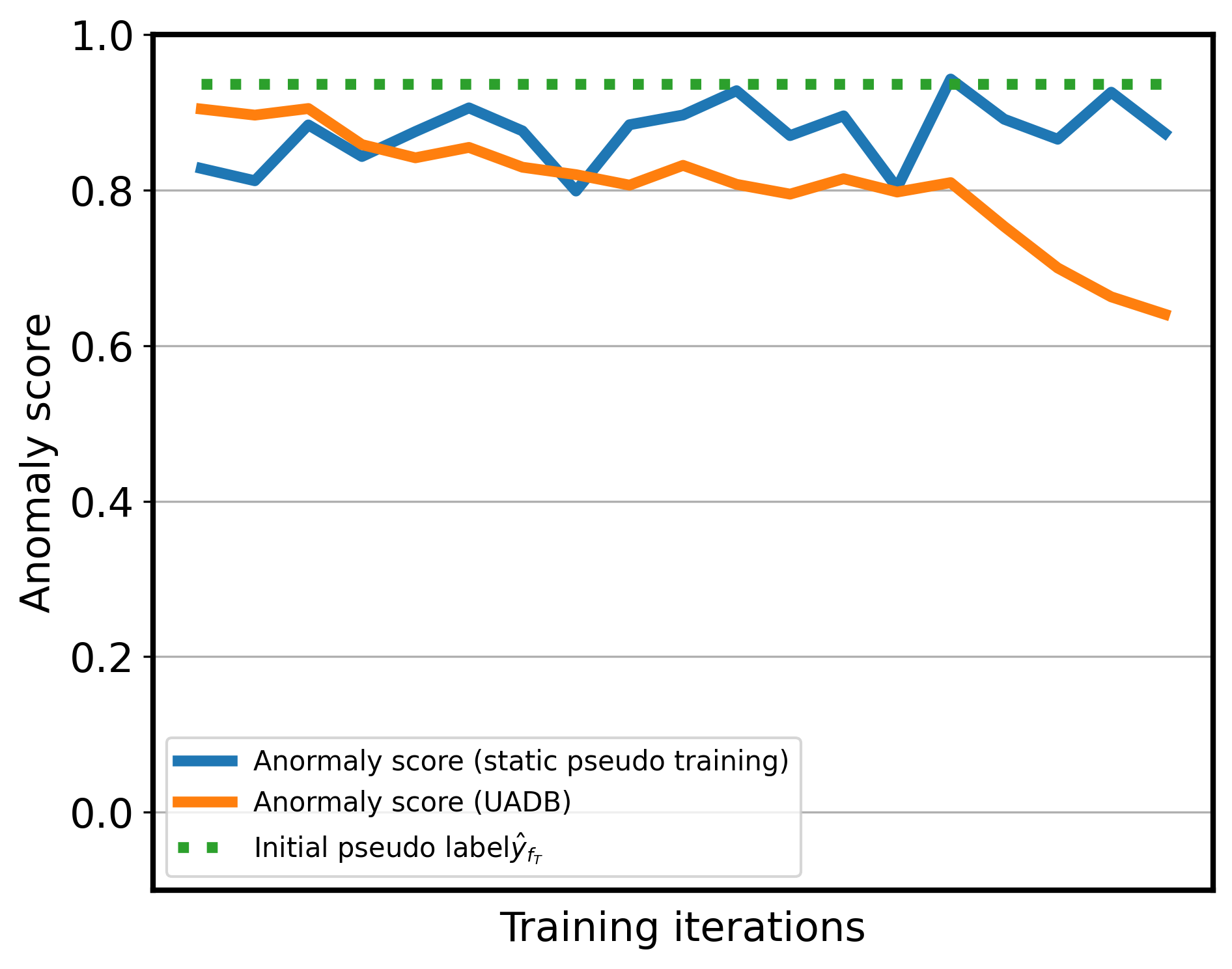}
\caption{False Positive (FP)}
\label{fig:01}
\end{subfigure}
\begin{subfigure}[b]{0.475\linewidth}
\centering
\includegraphics[width=\textwidth]{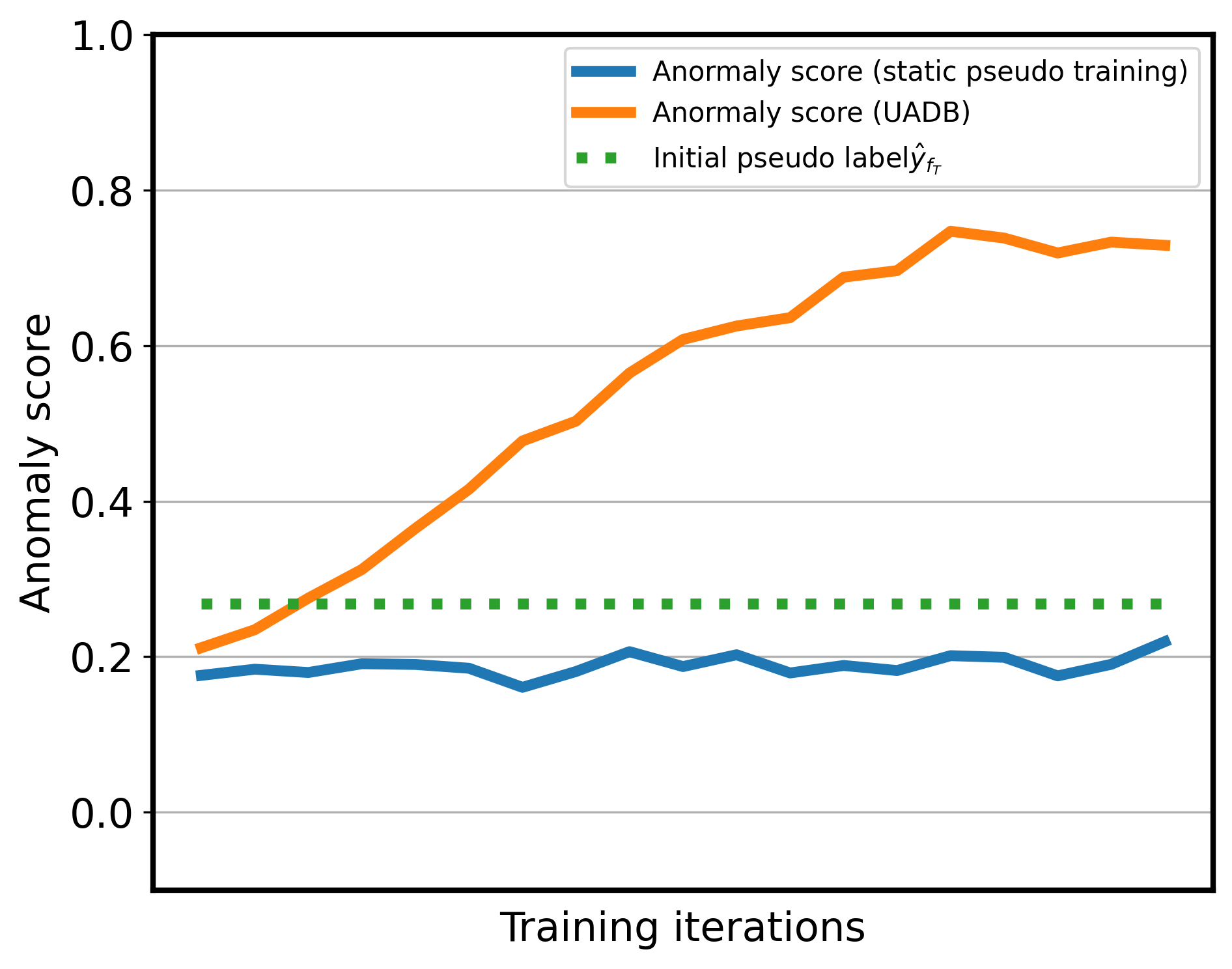}
\caption{False Negative (FN)}
\label{fig:10}
\end{subfigure}
\caption{
Error correction during the UADB training process.
    We compare UADB's behavior (orange line) with a variant that learns a student model by static pseudo-supervised training without error correction (blue line).
    The dashed line indicates the initial pseudo label $\hat{y}_{f_T}\in[0,1]$.
    We can observe that the student model without error correction simply mimics the teacher model's behavior (including errors), while UADB can gradually correct the booster's predictions on FP (Fig.~\ref{fig:01}) and FN (Fig.~\ref{fig:10}) by exploiting the variance difference between normal and abnormal instances.
}
\label{fig:correction}
\end{figure}

To validate previous analysis, for each case, we show an example of how the booster's prediction changes during the UADB training process.
Please see details in Fig.~\ref{fig:correction}.


\section{Experiments \& Analysis}
\label{sec:experiment}

In this section, we conduct comprehensive experiments to validate the effectiveness of the proposed \method framework.
We first introduce the experiment setup and the included 14 mainstream UAD models and 84 real-world datasets.
To provide an intuitive understanding of UADB, we construct synthetic datasets with different types of anomalies, and visualize the behaviors of example UAD models and their boosters.
After that, we test UADB on all 84 real-world tabular datasets and present the empirical results and corresponding analysis.
Finally, we conduct ablation study and further compare UADB with several intuitive booster frameworks.

\subsection{Experiment Setup Details.}
\label{sec:setup}
\textbf{Source UAD Models.}
\label{sec:mainstream UAD}
As the UADB is a model-agnostic framework, we include 14 mainstream UAD techniques to test UADB's applicability and effectiveness to different UAD models with diverse architecture.
These UAD algorithms include IForest\cite{liu2008isolation}, HBOS\cite{goldstein2012histogram}, LOF\cite{breunig2000lof}, KNN\cite{ramaswamy2000efficient}, PCA\cite{shyu2003novel}, OCSVM\cite{scholkopf1999support}, CBLOF\cite{he2003discovering}, COF\cite{tang2002enhancing}, SOD\cite{kriegel2009outlier}, ECOD\cite{li2022ecod}, GMM\cite{reynolds2009gaussian}, LODA\cite{pevny2016loda}, COPOD\cite{li2020copod} and DeepSVDD\cite{ruff2018deep}.
More details could be found in UADB's Github repository\footnote{\url{https://github.com/HangtingYe/UADB}}.

The aforementioned 14 UAD methods are widely used in practice and vary significantly in terms of both methodologies (e.g., neighbor-based~\cite{breunig2000lof,tang2002enhancing,ramaswamy2000efficient}, clustering-based~\cite{reynolds2009gaussian,he2003discovering}, density-based~\cite{goldstein2012histogram,li2022ecod}) and model architectures (e.g., tree~\cite{liu2008isolation}, support vector machine~\cite{scholkopf1999support}, neural network~\cite{ruff2018deep}).
Thus they can be used to perform a comprehensive test of the effectiveness and applicability of UADB.

\textbf{Real-world Datasets.}
As described before, UAD on tabular datasets are challenging due to the heterogeneity, complexity, and diversity of tabular data.
Table~\ref{tab:datasets} shows the statistics of the 84 heterogeneous tabular datasets that are included for a comprehensive evaluation.
Note that, in addition to the native tabular datasets (i.e., from abalone to yeast), datasets from larger and high-dimensional CV and NLP tasks are also included.
However, as many UAD models such as IForest \cite{liu2008isolation} and OCSVM \cite{scholkopf1999support} cannot directly handle CV or NLP task, we follow previous work \cite{han2022adbench} and use a CV/NLP feature extractor to generate tabular versions of these datasets.
These datasets vary significantly in properties and application domains, including health care (e.g., disease diagnosis), finance (e.g. credit card fraud detection), image processing (e.g. object identification), language processing (e.g. speech recognition) and more.

\textbf{UADB Setup.}
\label{sec:UADB setup}
As mentioned before, UADB adopt a neural network as the booster model for its strong expressive power as a universal approximator and ability to perform flexible post-hoc tuning.
We fix the parameter of source UAD model, and only optimize and keep the booster $f_B(\cdot;\theta)$ as the final UAD model.
Specifically, the booster model is a simple 3-layer fully-connected MLP (Multi-layer Perceptron) with 128 neurons in each hidden layer.
In $t$-th training step/iteration of UADB, the booster $f_B(\cdot;\theta)$ is updated w.r.t objective $\argmin_\theta\mathcal{L}(f_B(\textbf{X};\theta), \hat{\textbf{y}}^{(t)})$ for 10 epochs with batch size set to 256,
optimized by Adam optimizer with a learning rate of 0.001, and the total number of UADB training steps $T=10$.
In addition, to prevent the booster model from overfitting the source model, we train 3 booster models in a 3-fold cross-validation manner (i.e., each model was trained on different 2 out of 3 splits of data and pseudo-labels).
At inference time, we average the outputs of the 3 booster models as the final predictions.
To reduce the effect of randomness, the reported performance is averaged over 10 independent runs.

\begin{figure*}[t]
\centering
\includegraphics[width=0.836\linewidth]{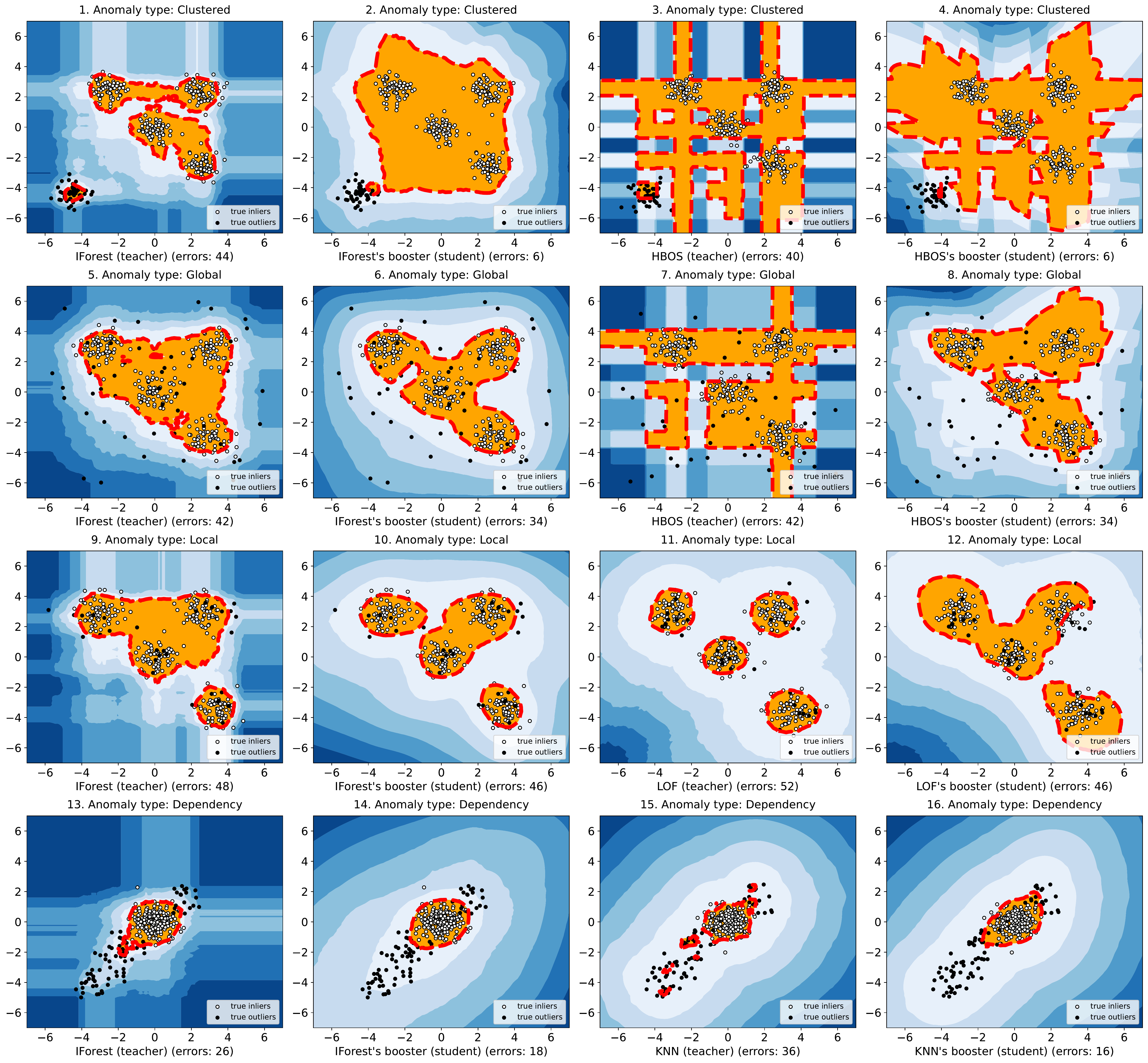}
\caption{
Visualization of the detection results of UAD models and their UADB boosters when facing different types of anomalies.
Specifically, we generate 4 synthetic datasets (in 4 rows) with different types of anomalies, i.e. Clustered, Global, Local and Dependency.
For each dataset, we test two UAD models (1st \& 3rd columns) that are best in handling the corresponding anomaly~\cite{han2022adbench} and their boosters (2nd \& 4th columns), we then plot their decision boundaries.
In each figure, the orange/blue interface represents the normal/abnormal space predicted by the model.
We can observe that UADB boosters can maintain the right decisions of the source model by knowledge transfer, while correcting the wrong predictions by adaptive error correction.
UADB improves the UAD model's detection results on all 8 model-anomaly-type pairs, achieving up to 86\% correction rate.
}
\label{fig:decision_boundary}
\end{figure*}

\textbf{Evaluation Metric \& Implementation Details.}
Following the common practice of previous works, we use \textit{Area Under the Curve of Receiver Characteristic Operator} (AUCROC) and \textit{Average Precision} (AP) to evaluate UAD models' performance.
Larger AUCROC/AP score indicates better detection performance.
Following ~\cite{han2022adbench}, we use the popular PyOD Python package~\cite{zhao2019pyod} to implement all source UAD models, and apply their default parameter settings in PyOD\footnote{Please refer to \url{https://pyod.readthedocs.io/en/latest/pyod.models.html}.}.
The booster model is implemented using the PyTorch~\cite{paszke2019pytorch} framework.

\begin{table*}[t]
\scriptsize
\centering
\caption{Data description of the 84 real-world datasets.}
\label{tab:datasets}
\setlength{\tabcolsep}{0.4mm}{
\begin{tabular}{ccc|ccc|ccc|ccc} 
\toprule
\textbf{Datasets}         & \textbf{\% Anomaly} & \textbf{Category} & \textbf{Datasets}    & \textbf{\% Anomaly} & \textbf{Category} & \textbf{Datasets}        & \textbf{\% Anomaly} & \textbf{Category} & \textbf{Datasets}        & \textbf{\% Anomaly} & \textbf{Category}  \\ 
\midrule
\textbf{abalone}          & 49.82               & Biology           & \textbf{mammography} & 2.32                & Healthcare        & \textbf{WDBC}            & 2.72                & Healthcare        & \textbf{FashionMNIST\_6} & 5.00                & Image              \\
\textbf{ALOI}             & 3.04                & Image             & \textbf{mnist}       & 9.21                & Image             & \textbf{Wilt}            & 5.33                & Botany            & \textbf{FashionMNIST\_7} & 5.00                & Image              \\
\textbf{annthyroid}       & 7.42                & Healthcare        & \textbf{musk}        & 3.17                & Chemistry         & \textbf{wine}            & 7.75                & Chemistry         & \textbf{FashionMNIST\_8} & 5.00                & Image              \\
\textbf{Arrhythmia}       & 45.78               & Healthcare        & \textbf{optdigits}   & 2.88                & Image             & \textbf{WPBC}            & 23.74               & Healthcare        & \textbf{FashionMNIST\_9} & 5.00                & Image              \\
\textbf{breastw}          & 34.99               & Healthcare        & \textbf{PageBlocks}  & 9.46                & Document          & \textbf{yeast}           & 34.16               & Biology           & \textbf{SVHN\_0}         & 5.00                & Image              \\
\textbf{cardio}           & 9.61                & Healthcare        & \textbf{Parkinson}   & 75.38               & Healthcare        & \textbf{CIFAR10\_0}      & 5.00                & Image             & \textbf{SVHN\_1}         & 5.00                & Image              \\
\textbf{Cardiotocography} & 22.04               & Healthcare        & \textbf{pendigits}   & 2.27                & Image             & \textbf{CIFAR10\_1}      & 5.00                & Image             & \textbf{SVHN\_2}         & 5.00                & Image              \\
\textbf{concrete}         & 50.00               & Physical          & \textbf{Pima}        & 34.90               & Healthcare        & \textbf{CIFAR10\_2}      & 5.00                & Image             & \textbf{SVHN\_3}         & 5.00                & Image              \\
\textbf{cover}            & 0.96                & Botany            & \textbf{satellite}   & 31.64               & Astronautics      & \textbf{CIFAR10\_3}      & 5.00                & Image             & \textbf{SVHN\_4}         & 5.00                & Image              \\
\textbf{fault}            & 34.67               & Physical          & \textbf{satimage-2}  & 1.22                & Astronautics      & \textbf{CIFAR10\_4}      & 5.00                & Image             & \textbf{SVHN\_5}         & 5.00                & Image              \\
\textbf{glass}            & 4.21                & Forensic          & \textbf{shuttle}     & 7.15                & Astronautics      & \textbf{CIFAR10\_5}      & 5.00                & Image             & \textbf{SVHN\_6}         & 5.00                & Image              \\
\textbf{HeartDisease}     & 44.44               & Healthcare        & \textbf{skin}        & 20.75               & Image             & \textbf{CIFAR10\_6}      & 5.00                & Image             & \textbf{SVHN\_7}         & 5.00                & Image              \\
\textbf{Hepatitis}        & 16.25               & Healthcare        & \textbf{smtp}        & 0.03                & Web               & \textbf{CIFAR10\_7}      & 5.00                & Image             & \textbf{SVHN\_8}         & 5.00                & Image              \\
\textbf{http}             & 0.39                & Web               & \textbf{SpamBase}    & 39.91               & Document          & \textbf{CIFAR10\_8}      & 5.00                & Image             & \textbf{SVHN\_9}         & 5.00                & Image              \\
\textbf{imgseg}           & 42.86               & image             & \textbf{speech}      & 1.65                & Linguistics       & \textbf{CIFAR10\_9}      & 5.00                & Image             & \textbf{agnews\_0}       & 5.00                & NLP                \\
\textbf{InternetAds}      & 18.72               & Image             & \textbf{Stamps}      & 9.12                & Document          & \textbf{FashionMNIST\_0} & 5.00                & Image             & \textbf{agnews\_1}       & 5.00                & NLP                \\
\textbf{Ionosphere}       & 35.90               & Oryctognosy       & \textbf{thyroid}     & 2.47                & Healthcare        & \textbf{FashionMNIST\_1} & 5.00                & Image             & \textbf{agnews\_2}       & 5.00                & NLP                \\
\textbf{landsat}          & 20.71               & Astronautics      & \textbf{vertebral}   & 12.50               & Biology           & \textbf{FashionMNIST\_2} & 5.00                & Image             & \textbf{agnews\_3}       & 5.00                & NLP                \\
\textbf{letter}           & 6.25                & Image             & \textbf{vowels}      & 3.43                & Linguistics       & \textbf{FashionMNIST\_3} & 5.00                & Image             & \textbf{amazon}          & 5.00                & NLP                \\
\textbf{Lymphography}     & 4.05                & Healthcare        & \textbf{Waveform}    & 2.90                & Physics           & \textbf{FashionMNIST\_4} & 5.00                & Image             & \textbf{imdb}            & 5.00                & NLP                \\
\textbf{magic.gamma}      & 35.16               & Physical          & \textbf{WBC}         & 4.48                & Healthcare        & \textbf{FashionMNIST\_5} & 5.00                & Image             & \textbf{yelp}            & 5.00                & NLP                \\
\bottomrule
\end{tabular}}
\end{table*}

\begin{table*}[t!]
\footnotesize
\centering
\caption{
The detection performance improvement achieved by UADB over 14 source UAD models on 84 datasets.
``\textbf{Original}'' indicates the average score achieved by the corresponding source UAD model on all datasets.
``\textbf{Improvement}'' indicates the average score improvement achieved by the UADB booster over the UAD model, ``\textbf{Improvement (\%)}'' indicates the average score improvement in percentage.
``\textbf{Effects}'' represents the number of datasets that UADB booster made improvements over the source model.
``\textbf{P-value}'' represents the results of the Wilcoxon signed-rank test (with $\alpha=0.05$), ``\textbf{P-value}'' less than 0.05 indicates the improvement is statistically significant.
}
\label{tab:results_statistic}
\setlength{\tabcolsep}{0.5mm}{
\begin{tabular}{c|c|cccccccccccccc} 
\toprule
\multicolumn{2}{c|}{\textbf{Source UAD Model}}                                          & \textbf{IForest}                   & \textbf{\textbf{HBOS}}             & \textbf{LOF}                        & \textbf{KNN}                        & \textbf{\textbf{PCA}}              & \textbf{OCSVM}                     & \textbf{CBLOF}                     & \textbf{COF}                        & \textbf{SOD}                        & \textbf{ECOD}                      & \textbf{\textbf{GMM}}               & \textbf{LODA}                       & \textbf{COPOD}                     & \textbf{\textbf{DeepSVDD}}           \\ 
\midrule
\multirow{5}{*}{\textbf{AUCROC}} & \textbf{Original}                                    & 0.7028                             & 0.6848                             & 0.6311                              & 0.6794                              & 0.6930                             & 0.6750                             & 0.7110                             & 0.6105                              & 0.6638                              & 0.6866                             & 0.7274                              & 0.6571                              & 0.6882                             & 0.5346                               \\ 
\cmidrule{2-16}
                                 & \textbf{Improvement}                                 & 0.0117                             & 0.0153                             & 0.0694                              & 0.0452                              & 0.0130                             & 0.0241                             & 0.0215                             & 0.0580                              & 0.0562                              & 0.0144                             & 0.0133                              & 0.0293                              & 0.0116                             & 0.0979                               \\ 
\cmidrule{2-16}
                                 & \textbf{\textbf{Improvement (\%)}}                   & 1.66                               & \textcolor[rgb]{0.2,0.2,0.2}{2.23} & 11.00                               & \textcolor[rgb]{0.2,0.2,0.2}{6.65}  & \textcolor[rgb]{0.2,0.2,0.2}{1.88} & \textcolor[rgb]{0.2,0.2,0.2}{3.57} & \textcolor[rgb]{0.2,0.2,0.2}{3.02} & \textcolor[rgb]{0.2,0.2,0.2}{9.50}  & \textcolor[rgb]{0.2,0.2,0.2}{8.46}  & \textcolor[rgb]{0.2,0.2,0.2}{2.10} & \textcolor[rgb]{0.2,0.2,0.2}{1.83}  & \textcolor[rgb]{0.2,0.2,0.2}{4.46}  & \textcolor[rgb]{0.2,0.2,0.2}{1.69} & \textcolor[rgb]{0.2,0.2,0.2}{18.31}  \\ 
\cmidrule{2-16}
                                 & \textbf{Effects}                                     & 49                                 & 59                                 & 51                                  & 46                                  & 53                                 & 58                                 & 51                                 & 57                                  & 57                                  & 57                                 & 47                                  & 62                                  & 52                                 & 68                                   \\ 
\cmidrule{2-16}
                                 & \textbf{P-value}                                     & 1.89e-2                            & 1.99e-4                            & 7.18e-4                             & 8.51e-4                             & 1.48e-3                            & 2.61e-6                            & 3.06e-3                            & 5.71e-5                             & 3.19e-6                             & 1.31e-4                            & 1.80e-2                             & 4.62e-6                             & 1.20e-2                            & 3.53e-11                             \\ 
\midrule
\multirow{5}{*}{\textbf{AP }}    & \textbf{Original}                                    & 0.3012                             & 0.2918                             & 0.1903                              & 0.2550                              & 0.3051                             & 0.2738                             & 0.3057                             & 0.1989                              & 0.2322                              & 0.2908                             & 0.2805                              & 0.2636                              & 0.2832                             & 0.1727                               \\ 
\cmidrule{2-16}
                                 & \textbf{Improvement}                                 & 0.0134                             & 0.0137                             & 0.1146                              & 0.0627                              & 0.0010                             & 0.0229                             & 0.0184                             & 0.0670                              & 0.0742                              & 0.0101                             & 0.0283                              & 0.0390                              & 0.0146                             & 0.0741                               \\ 
\cmidrule{2-16}
                                 & \textbf{\textbf{\textbf{\textbf{Improvement (\%)}}}} & \textcolor[rgb]{0.2,0.2,0.2}{4.45} & \textcolor[rgb]{0.2,0.2,0.2}{4.69} & \textcolor[rgb]{0.2,0.2,0.2}{60.22} & \textcolor[rgb]{0.2,0.2,0.2}{24.59} & \textcolor[rgb]{0.2,0.2,0.2}{0.32} & \textcolor[rgb]{0.2,0.2,0.2}{8.36} & \textcolor[rgb]{0.2,0.2,0.2}{6.02} & \textcolor[rgb]{0.2,0.2,0.2}{33.69} & \textcolor[rgb]{0.2,0.2,0.2}{31.96} & \textcolor[rgb]{0.2,0.2,0.2}{3.47} & \textcolor[rgb]{0.2,0.2,0.2}{10.09} & \textcolor[rgb]{0.2,0.2,0.2}{14.80} & \textcolor[rgb]{0.2,0.2,0.2}{5.16} & \textcolor[rgb]{0.2,0.2,0.2}{42.91}  \\ 
\cmidrule{2-16}
                                 & \textbf{Effects}                                     & 57                                 & 62                                 & 56                                  & 51                                  & 62                                 & 59                                 & 57                                 & 60                                  & 60                                  & 66                                 & 51                                  & 64                                  & 64                                 & 70                                   \\ 
\cmidrule{2-16}
                                 & \textbf{P-value}                                     & 1.77e-4                            & 2.00e-6                            & 1.21e-6                             & 3.50e-6                             & 3.64e-6                            & 1.20e-8                            & 1.96e-4                            & 5.22e-7                             & 7.23e-9                             & 2.29e-8                            & 1.26e-3                             & 4.56e-7                             & 2.38e-7                            & 1.74e-10                             \\
\bottomrule
\end{tabular}}
\end{table*}

\subsection{Visualization on Synthetic Datasets}
\label{sec:main results}
Before diving into the results on real-world datasets, we first show some visualizations on synthetic datasets to provide an intuitive understanding of how UADB works and improves over the original model.
Previous research \cite{han2022adbench, gopalan2019pidforest} have demonstrated that anomalies in real-world applications could be roughly divided into four specific types, i.e. clustered, global, local, and dependency anomalies.
Accordingly, we generate 4 synthetic datasets, each contains a specific type of anomaly, as shown in the rows of Fig.~\ref{fig:decision_boundary}.
For each type of anomaly, we select two UAD models that generally perform best on the given anomaly, then apply UADB to get two booster models, and show their predictions and errors in Fig.~\ref{fig:decision_boundary}.

It can be observed that UADB generally maintains the predictions of the teacher model by knowledge transfer.
But more importantly, the booster models benefit from the adaptive correction mechanism and thus are able to correct the errors when learning from the teacher model, e.g., the false positives in IForest-Clustered (1th-row left) and the false negatives in HBOS-Global (2nd-row right).
On average, UADB achieves 38.94\% error correction rate on all 8 Model-Anomaly pairs, with a maximum at 86.36\% on IForest-Clustered (1st-row left), where 38 out of 44 errors of the source IForest model are corrected in its booster.
These results show that UADB is able to handle different type of anomalies, and can make improvements even over the best-performing UAD models.

\begin{table*}
\scriptsize
\centering
\caption{
\method's performance on representative UAD models (teacher) in terms of AUCROC and AP (higher is better).
Due to the space limitations, we select 4 widely used UAD techniques (i.e., IForest~\cite{liu2008isolation}, HBOS~\cite{goldstein2012histogram}, LOF~\cite{breunig2000lof}, and KNN~\cite{ramaswamy2000efficient}) as representatives and show the performance of them and their UADB boosters on 5 example datasets.
In each sub-table, we show the teacher model's performance as well as the booster's performance (during and after UADB training) on the 5 datasets.
The ``improvement'' indicates the performance improvement that the UADB booster achieved over its source UAD model.
In each sub-table row, we show the performance sub-table in terms of AUCROC and AP for a specific UAD model.
}
\label{tab:results with multiple iterations}
\setlength{\tabcolsep}{0.8mm}{
\begin{tabular}{c|c|ccccc|ccc|c|ccccc|c} 
\cmidrule[\heavyrulewidth]{1-8}\cmidrule[\heavyrulewidth]{10-17}
\multicolumn{8}{c}{\textbf{(a)} Performance of (\textbf{IForest}) and its UADB booster in terms of \textbf{AUCROC}}                                                                  &  & \multicolumn{8}{c}{\textbf{(b)} Performance of (\textbf{IForest}) and its UADB booster in terms of \textbf{AP}}                                                                       \\ 
\cmidrule{1-8}\cmidrule{10-17}
\textbf{Datasets}        & \textbf{Teacher}     & \textbf{iter 2} & \textbf{iter 4} & \textbf{iter~6} & \textbf{iter~8} & \textbf{iter 10}     & \textbf{Improvement} &  & \textbf{Datasets}        & \textbf{Teacher}     & \textbf{iter 2} & \textbf{iter~4} & \textbf{iter~6} & \textbf{iter~8} & \textbf{iter~10}     & \textbf{Improvement}  \\ 
\cmidrule{1-8}\cmidrule{10-17}
\textbf{speech}          & 0.5057               & 0.579           & 0.6002          & 0.613           & 0.62            & 0.6233               & 0.1176               &  & \textbf{pendigits}       & 0.3392               & 0.3995          & 0.4154          & 0.4505          & 0.4973          & 0.5524               & 0.2132                \\
\textbf{Wilt}            & 0.4276               & 0.4407          & 0.4989          & 0.5305          & 0.5309          & 0.5364               & 0.1088               &  & \textbf{vowels}          & 0.1825               & 0.1743          & 0.1835          & 0.2463          & 0.3143          & 0.3408               & 0.1582                \\
\textbf{satellite}       & 0.6668               & 0.6988          & 0.7117          & 0.725           & 0.7413          & 0.7625               & 0.0957               &  & \textbf{satellite}       & 0.6248               & 0.7074          & 0.7163          & 0.7252          & 0.7317          & 0.7399               & 0.1151                \\
\textbf{vowels}          & 0.8118               & 0.8318          & 0.8488          & 0.8707          & 0.8918          & 0.9066               & 0.0949               &  & \textbf{InternetAds}     & 0.5078               & 0.5166          & 0.5221          & 0.5313          & 0.5468          & 0.5588               & 0.051                 \\
\textbf{abalone}         & 0.4989               & 0.5435          & 0.5524          & 0.5589          & 0.5603          & 0.5663               & 0.0674               &  & \textbf{abalone}         & 0.5111               & 0.531           & 0.5379          & 0.5443          & 0.5468          & 0.5529               & 0.0418                \\
\cmidrule[\heavyrulewidth]{1-8}\cmidrule[\heavyrulewidth]{10-17}
\multicolumn{1}{c}{}     & \multicolumn{1}{c}{} &                 &                 &                 &                 & \multicolumn{1}{c}{} &                      &  & \multicolumn{1}{c}{}     & \multicolumn{1}{c}{} &                 &                 &                 &                 & \multicolumn{1}{c}{} &                       \\ 
\cmidrule[\heavyrulewidth]{1-8}\cmidrule[\heavyrulewidth]{10-17}
\multicolumn{8}{c}{\textbf{(c)} Performance of (\textbf{HBOS}) and its UADB booster in terms of \textbf{AUCROC}}                                                                     &  & \multicolumn{8}{c}{\textbf{(d)} Performance of (\textbf{HBOS}) and its UADB booster in terms of \textbf{AP}}                                                                          \\ 
\cmidrule{1-8}\cmidrule{10-17}
\textbf{Datasets}        & \textbf{Teacher}     & \textbf{iter 2} & \textbf{iter 4} & \textbf{iter~6} & \textbf{iter~8} & \textbf{iter 10}     & \textbf{Improvement} &  & \textbf{Datasets}        & \textbf{Teacher}     & \textbf{iter 2} & \textbf{iter~4} & \textbf{iter~6} & \textbf{iter~8} & \textbf{iter~10}     & \textbf{Improvement}  \\ 
\cmidrule{1-8}\cmidrule{10-17}
\textbf{speech}          & 0.4673               & 0.5931          & 0.612           & 0.626           & 0.6299          & 0.633                & 0.1657               &  & \textbf{pendigits}       & 0.2805               & 0.3397          & 0.359           & 0.3671          & 0.3779          & 0.389                & 0.1085                \\
\textbf{Wilt}            & 0.3849               & 0.5111          & 0.5311          & 0.5492          & 0.5464          & 0.5426               & 0.1577               &  & \textbf{vowels}          & 0.0883               & 0.0865          & 0.1052          & 0.1254          & 0.1524          & 0.182                & 0.0936                \\
\textbf{vowels}          & 0.6726               & 0.7341          & 0.7732          & 0.7878          & 0.7987          & 0.8051               & 0.1325               &  & \textbf{Ionosphere}      & 0.4133               & 0.4516          & 0.5228          & 0.5313          & 0.5095          & 0.486                & 0.0727                \\
\textbf{mnist}           & 0.6141               & 0.6662          & 0.6623          & 0.6672          & 0.675           & 0.6779               & 0.0638               &  & \textbf{imgseg}          & 0.6105               & 0.644           & 0.6478          & 0.6476          & 0.6509          & 0.6538               & 0.0433                \\
\textbf{imgseg}          & 0.656                & 0.6945          & 0.7012          & 0.7032          & 0.7058          & 0.7096               & 0.0536               &  & \textbf{cardio}          & 0.4679               & 0.5132          & 0.5182          & 0.5134          & 0.5099          & 0.508                & 0.0401                \\
\cmidrule[\heavyrulewidth]{1-8}\cmidrule[\heavyrulewidth]{10-17}
\multicolumn{1}{c}{}     & \multicolumn{1}{c}{} &                 &                 &                 &                 & \multicolumn{1}{c}{} &                      &  & \multicolumn{1}{c}{}     & \multicolumn{1}{c}{} &                 &                 &                 &                 & \multicolumn{1}{c}{} &                       \\ 
\cmidrule[\heavyrulewidth]{1-8}\cmidrule[\heavyrulewidth]{10-17}
\multicolumn{8}{c}{\textbf{(e)} Performance of (\textbf{LOF}) and its UADB booster in terms of \textbf{AUCROC}}                                                                      &  & \multicolumn{8}{c}{\textbf{(f)} Performance of (\textbf{LOF}) and its UADB booster in terms of \textbf{AP}}                                                                           \\ 
\cmidrule{1-8}\cmidrule{10-17}
\textbf{Datasets}        & \textbf{Teacher}     & \textbf{iter 2} & \textbf{iter 4} & \textbf{iter~6} & \textbf{iter~8} & \textbf{iter 10}     & \textbf{Improvement} &  & \textbf{Datasets}        & \textbf{Teacher}     & \textbf{iter 2} & \textbf{iter~4} & \textbf{iter~6} & \textbf{iter~8} & \textbf{iter~10}     & \textbf{Improvement}  \\ 
\cmidrule{1-8}\cmidrule{10-17}
\textbf{http}            & 0.3685               & 1               & 1               & 1               & 1               & 1                    & 0.6315               &  & \textbf{http}            & 0.0603               & 1               & 1               & 1               & 1               & 1                    & 0.9397                \\
\textbf{shuttle}         & 0.4886               & 0.9199          & 0.9244          & 0.9691          & 0.9537          & 0.9525               & 0.4638               &  & \textbf{shuttle}         & 0.0958               & 0.6814          & 0.7598          & 0.873           & 0.7902          & 0.7882               & 0.6924                \\
\textbf{satimage-2}      & 0.4702               & 0.5677          & 0.6909          & 0.7618          & 0.8402          & 0.9146               & 0.4444               &  & \textbf{optdigits}       & 0.0732               & 0.1114          & 0.1825          & 0.2627          & 0.3593          & 0.4551               & 0.3819                \\
\textbf{optdigits}       & 0.5819               & 0.8488          & 0.8884          & 0.9128          & 0.9318          & 0.9438               & 0.3619               &  & \textbf{satellite}       & 0.3746               & 0.6268          & 0.675           & 0.6887          & 0.7047          & 0.7117               & 0.3371                \\
\textbf{musk}            & 0.4586               & 0.5008          & 0.5584          & 0.6137          & 0.6848          & 0.7528               & 0.2942               &  & \textbf{WDBC}            & 0.1026               & 0.1757          & 0.2627          & 0.3202          & 0.3604          & 0.3754               & 0.2727                \\
\cmidrule[\heavyrulewidth]{1-8}\cmidrule[\heavyrulewidth]{10-17}
\multicolumn{1}{c}{}     & \multicolumn{1}{c}{} &                 &                 &                 &                 & \multicolumn{1}{c}{} &                      &  & \multicolumn{1}{c}{}     & \multicolumn{1}{c}{} &                 &                 &                 &                 & \multicolumn{1}{c}{} &                       \\ 
\cmidrule[\heavyrulewidth]{1-8}\cmidrule[\heavyrulewidth]{10-17}
\multicolumn{8}{c}{\textbf{(g)} Performance of (\textbf{KNN}) and its UADB booster in terms of \textbf{AUCROC}}                                                                      &  & \multicolumn{8}{c}{\textbf{(h)} Performance of (\textbf{KNN}) and its UADB booster in terms of \textbf{AP}}                                                                           \\ 
\cmidrule{1-8}\cmidrule{10-17}
\textbf{Datasets}        & \textbf{Teacher}     & \textbf{iter 2} & \textbf{iter 4} & \textbf{iter~6} & \textbf{iter~8} & \textbf{iter 10}     & \textbf{Improvement} &  & \textbf{Datasets}        & \textbf{Teacher}     & \textbf{iter 2} & \textbf{iter~4} & \textbf{iter~6} & \textbf{iter~8} & \textbf{iter~10}     & \textbf{Improvement}  \\ 
\cmidrule{1-8}\cmidrule{10-17}
\textbf{Stamps}          & 0.6587               & 0.773           & 0.8458          & 0.8863          & 0.8876          & 0.8851               & 0.2264               &  & \textbf{shuttle}         & 0.1558               & 0.9055          & 0.9203          & 0.8229          & 0.7425          & 0.7379               & 0.5822                \\
\textbf{SpamBase}        & 0.5086               & 0.6249          & 0.6675          & 0.692           & 0.7087          & 0.7119               & 0.2033               &  & \textbf{satimage-2}      & 0.3294               & 0.4703          & 0.5803          & 0.6749          & 0.8082          & 0.8873               & 0.5579                \\
\textbf{pendigits}       & 0.7133               & 0.8409          & 0.8484          & 0.8691          & 0.8861          & 0.9047               & 0.1914               &  & \textbf{satellite}       & 0.5041               & 0.7098          & 0.7306          & 0.7448          & 0.7609          & 0.7683               & 0.2642                \\
\textbf{musk}            & 0.6845               & 0.775           & 0.7961          & 0.8153          & 0.8377          & 0.8551               & 0.1706               &  & \textbf{SpamBase}        & 0.3967               & 0.4643          & 0.5138          & 0.5479          & 0.5746          & 0.5815               & 0.1848                \\
\textbf{shuttle}         & 0.6618               & 0.9908          & 0.9903          & 0.9365          & 0.8425          & 0.8134               & 0.1516               &  & \textbf{WPBC}            & 0.232                & 0.2343          & 0.2788          & 0.3481          & 0.3952          & 0.4123               & 0.1803                \\
\cmidrule[\heavyrulewidth]{1-8}\cmidrule[\heavyrulewidth]{10-17}
\end{tabular}}
\end{table*}

\subsection{Results \& Analysis on Real-world Datasets}
\label{sec:main results}

In this section, we carry out extensive experiments on 84 heterogeneous real-world datasets with 14 different mainstream UAD models.
The results can provide a comprehensive evaluation of the effectiveness of UADB in real-world UAD tasks.
The statistics of the 84 heterogeneous tabular datasets are shown in Table~\ref{tab:datasets}.
Specifically, we aim to answer the following research questions (\textbf{RQ}s) in this section:
\begin{enumerate}
    \item \textbf{RQ1}: Can UADB generally boost the performance of different UAD models in real-world applications?
    \item \textbf{RQ2}: How does the error correction mechanism work? And what is the role of iterative training?
    \item \textbf{RQ3}: How much did the iterative error correction mechanism contribute to UADB training?
    \item \textbf{RQ4}: Can other intuitive mechanisms that exploit the variance also improve over UAD models?
\end{enumerate}

\textbf{Main Results (RQ1).}
First, we train the 14 different mainstream UAD models as well as their corresponding UADB boosters on all 84 tabular datasets.
Note that this is a large-scale experiment that result in 14(models)$\times$84(datasets)$\times$2(metrics)$\times$2(source \& booster) = 4704 numerical results.
We cannot fully display these results here due to space limitations.
Therefore, we provide a summarization of the experimental results in Table \ref{tab:results_statistic}, and part of the more detailed results are available in Table \ref{tab:results with multiple iterations}.

From Table \ref{tab:results_statistic} we can observe that:
\begin{itemize}
    \item
    UADB achieves consistent improvements over the 14 source UAD models for 84 tabular datasets. It achieved a more than 1\% average performance improvement for each model, whether using AUCROC or AP as the evaluation metric.
    In addition, we conduct Wilcoxon signed-rank test (with $\alpha=0.05$) for each source UAD model and its booster over 84 datasets.
    In all settings, the improvement is statistically significant at the 95\% confidence level according to the Wilcoxon signed-rank test.
    This demonstrates the superior adaptability and generality of UADB to different models and tasks.
    \item 
    There is no UAD model that consistently outperforms others on all tasks, i.e., the universal winner UAD solution does not exist. This is aligned with the findings in previous works~\cite{han2022adbench}.
    \item
    Using a different evaluation metric can lead to a different ``best model''. For example, GMM~\cite{reynolds2009gaussian} obtains the best average performance in terms of AUCROC, but CBLOF~\cite{he2003discovering} becomes the best performer if we change the evaluation metric to AP.
    \item
    UADB can make improvements even over the best performer UAD model (i.e., GMM~\cite{reynolds2009gaussian} in terms of AUCROC and CBLOF~\cite{he2003discovering} in terms of AP). Despite their relatively good performance, UADB still managed to achieve an average 1.83\%/10.09\% relative performance gain of AUCROC/AP over GMM, and average 3.02\%/6.02\% relative performance gain of AUCROC/AP over CBLOF.
\end{itemize}

\begin{figure}[t]
\centering
\includegraphics[width = \linewidth]{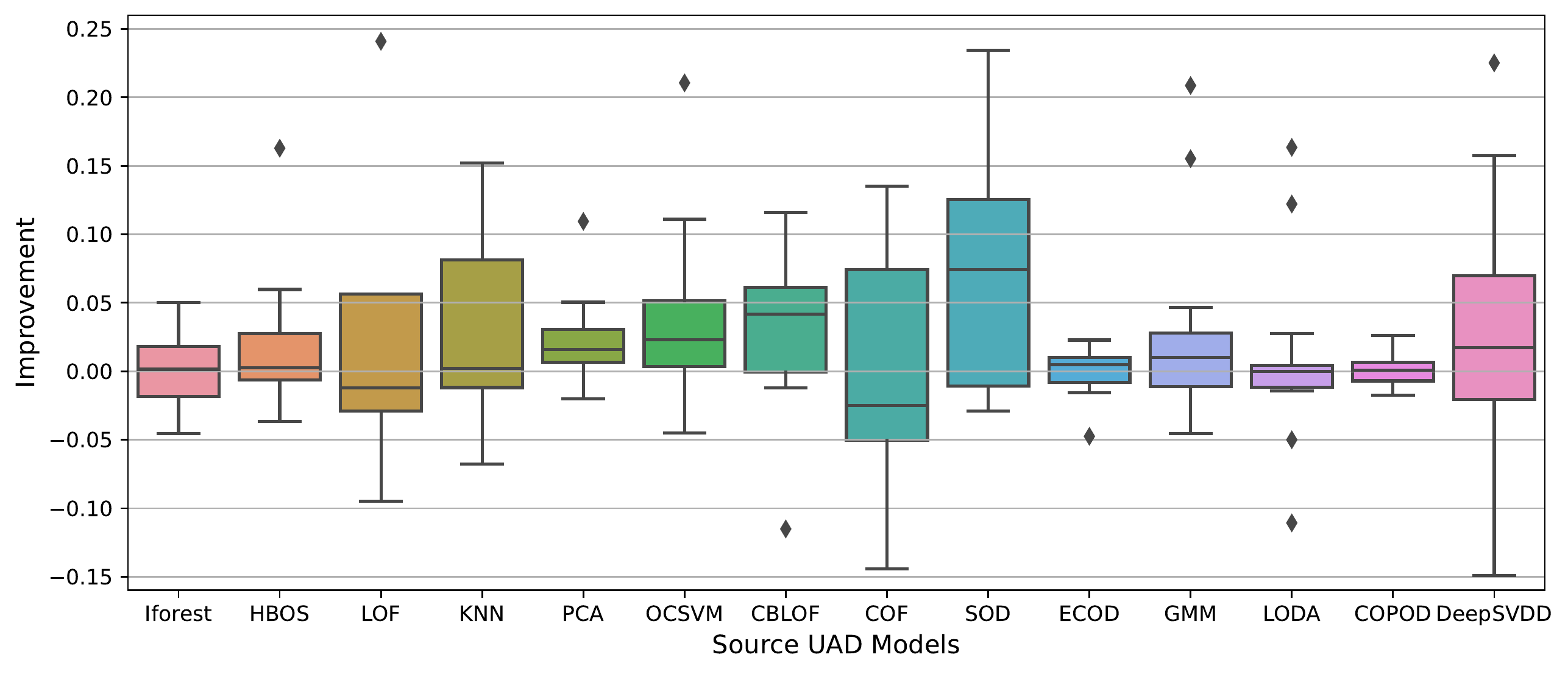}
\caption{UADB's performance on datasets that anomalies do not have higher average variance than normal samples.
“Improvement” indicates the improvements achieved by the UADB booster over the UAD model on these datasets.
}
\label{fig:case_unsatisfy}
\end{figure}

As observed in Fig.~\ref{fig:std_statistic}, for a small portion of datasets, anomalies do not have higher average variance than normal samples.
We further explore UADB's performance on these datasets, which are shown in Fig.~\ref{fig:case_unsatisfy}.
Though the empirical evidence does not hold, UADB could still achieve improvements over 12 out of 14 UAD models on more than half of these datasets.

\begin{figure}[h]
\centering
\includegraphics[width = 0.95\linewidth]{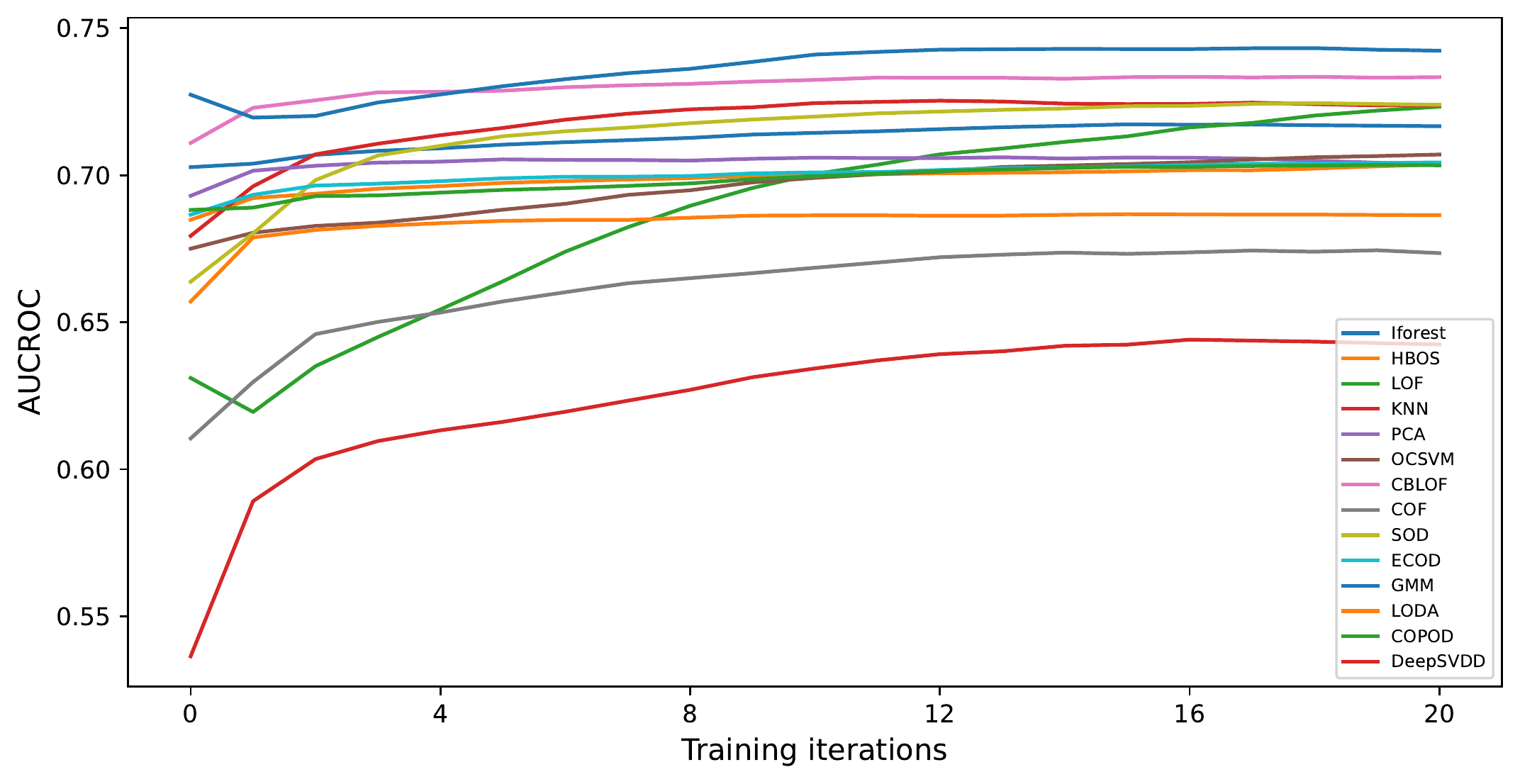}
\caption{UADB's performance (AUCROC) with different number of training iterations/steps.
The results are averaged over 84 tabular datasets.
}
\label{fig:performance_become_stable}
\vspace{-1em}
\end{figure}

\begin{figure}[h]
\centering
\includegraphics[width = 0.95\linewidth]{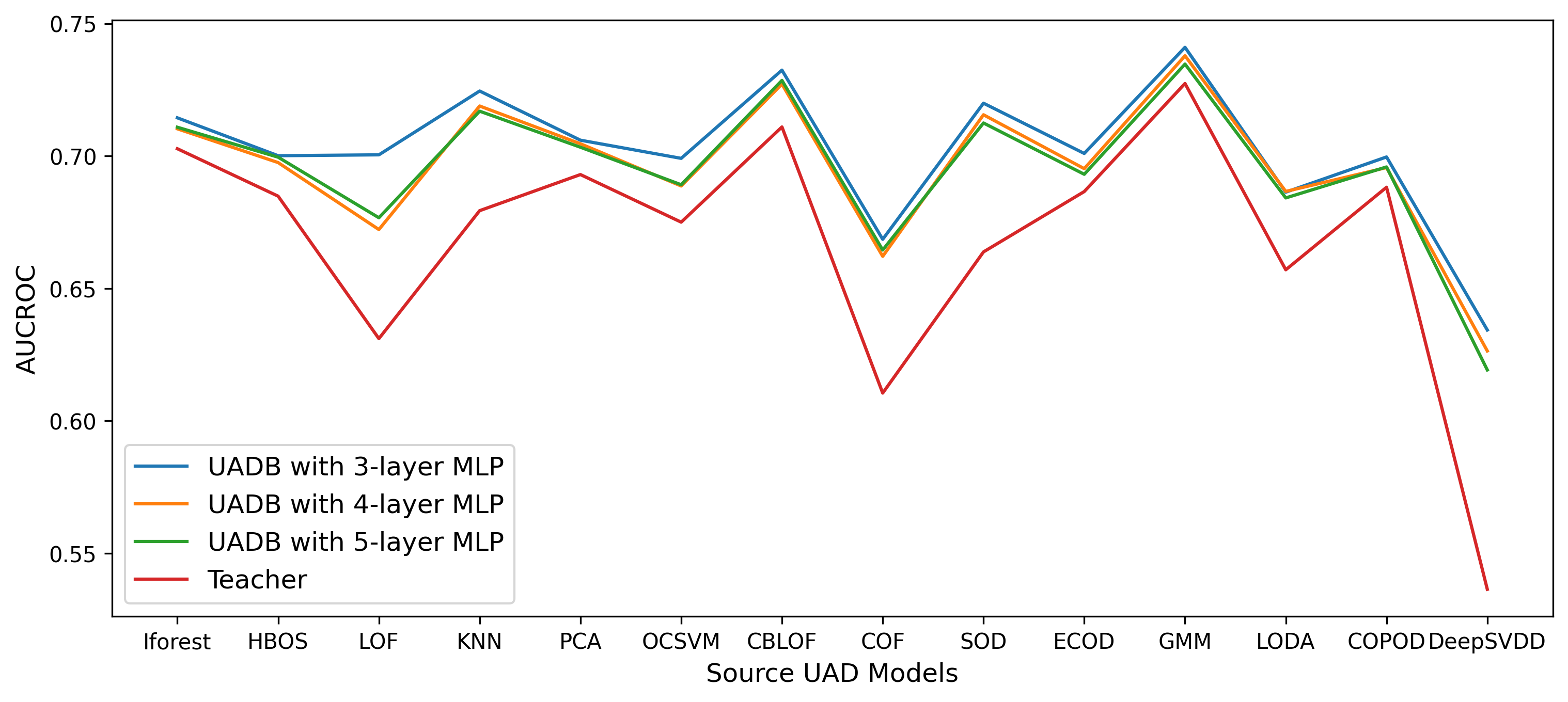}
\caption{UADB's performance (AUCROC) with different MLP layers.
The results are averaged over 84 tabular datasets.}
\label{fig:different layers}
\vspace{-1em}
\end{figure}

\textbf{Sensitivity Analysis.}
Fig.~\ref{fig:performance_become_stable} demonstrated that for most UAD models, the performance of the UADB gradually improves with an increase in training iterations, and the UADB's performance reaches a stable level after 10 training iterations.
Thus, it is reasonable to set the total training steps $T$ to 10 in Section ~\ref{sec:UADB setup}.
In addition, Fig.~\ref{fig:different layers} shows the AUCROC performance of UADB w.r.t. different MLP layers.
The results indicate that UADB performs very stably w.r.t. the number of MLP layers on all datasets. 

\begin{figure}[b]
\vspace{-1em}
\centering
\begin{subfigure}[b]{\linewidth}
\centering
\includegraphics[width=\textwidth]{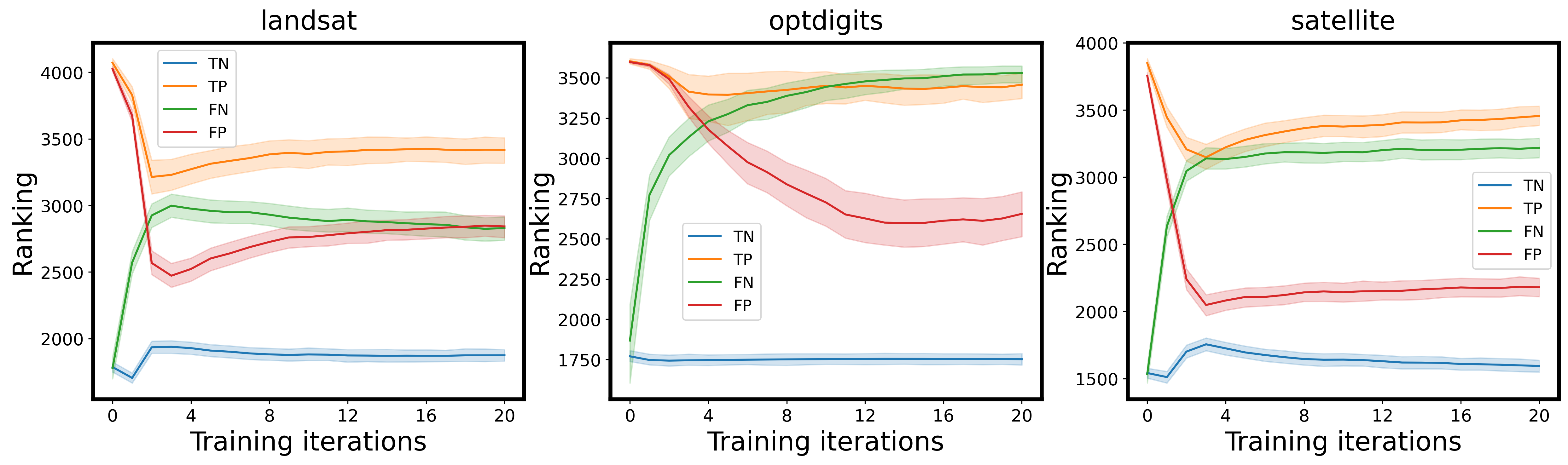}
\caption{The development of ranking.}
\label{fig:ranking}
\end{subfigure}
\begin{subfigure}[b]{\linewidth}
\centering
\includegraphics[width=\textwidth]{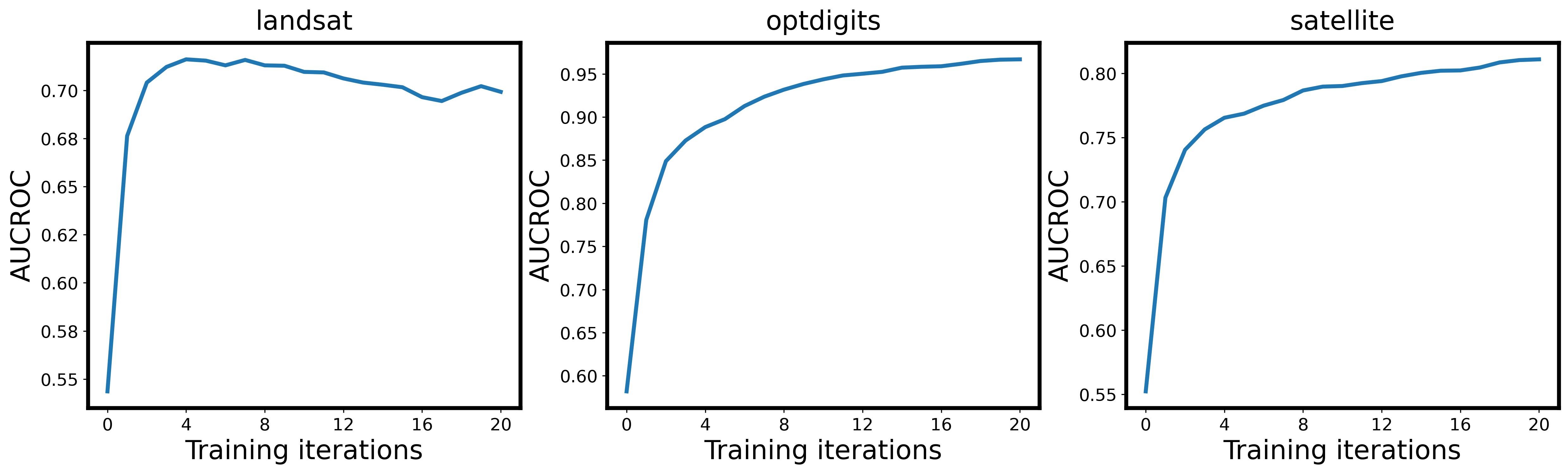}
\caption{The development of AUCROC performance.}
\label{fig:aucroc}
\end{subfigure}
\caption{
The development of instances' ranking and UADB's performance on \textit{landsat, optdigits and satellite}.
Here, we adopt LOF as UAD model and the number of training steps is set to 20.
We reported the average ranking of 4 types of instances (TP, TN, FP, FN) respectively.
}
\label{fig:case study on real world datasets}
\end{figure}

\label{sec:RQ2}
\textbf{Case Study on Real-world Datasets (RQ2).}
To further validate the role of iterative training, we show some real-world dataset cases for better understanding.
In Fig.~\ref{fig:case study on real world datasets}, we show the ranking development of 4 types of instances (i.e. True Positive (TP), True Negative (TN), False Positive (FP), and False Negative (FN)) on several real-world datasets, and also show the respective performance of UAD booster during multiple iterations.
Higher ranking represents higher anomaly score.
Let's first consider TP and FP, the initial pseudo labels (teacher's prediction) are closer to 1 (i.e. the ranking is high), however, the variance of TP is higher than FP (anomalies has higher average variance compared
to normal samples), thus after adding the variance to pseudo labels, the ranking of FP would decrease compared to TP.
After multiple iterations, the difference in ranking between FP and TP will increase (TP maintains high ranking while the ranking of FP decreases).
Likewise, the difference in ranking between FN and TN will also increase.
Therefore, after multiple training iterations, UADB could maintain the right decisions of the source model, while correcting the wrong predictions by error correction mechanism.

\begin{figure}[h]
\vspace{-1em}
\centering
\begin{subfigure}[b]{0.95\linewidth}
\centering
\includegraphics[width=\textwidth]{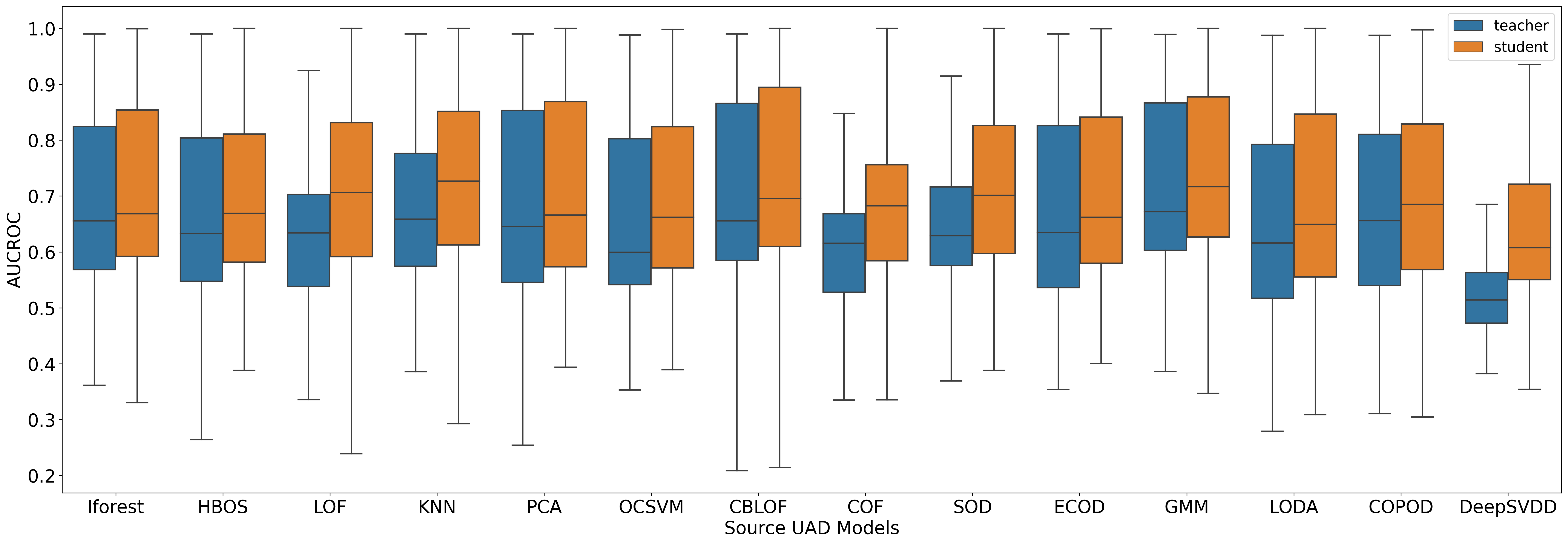}
\caption{Evaluation metric: Area Under the Curve of Receiver Characteristic Operator (AUCROC)}
\label{fig:y equals x}
\end{subfigure}
\begin{subfigure}[b]{0.95\linewidth}
\centering
\includegraphics[width=\textwidth]{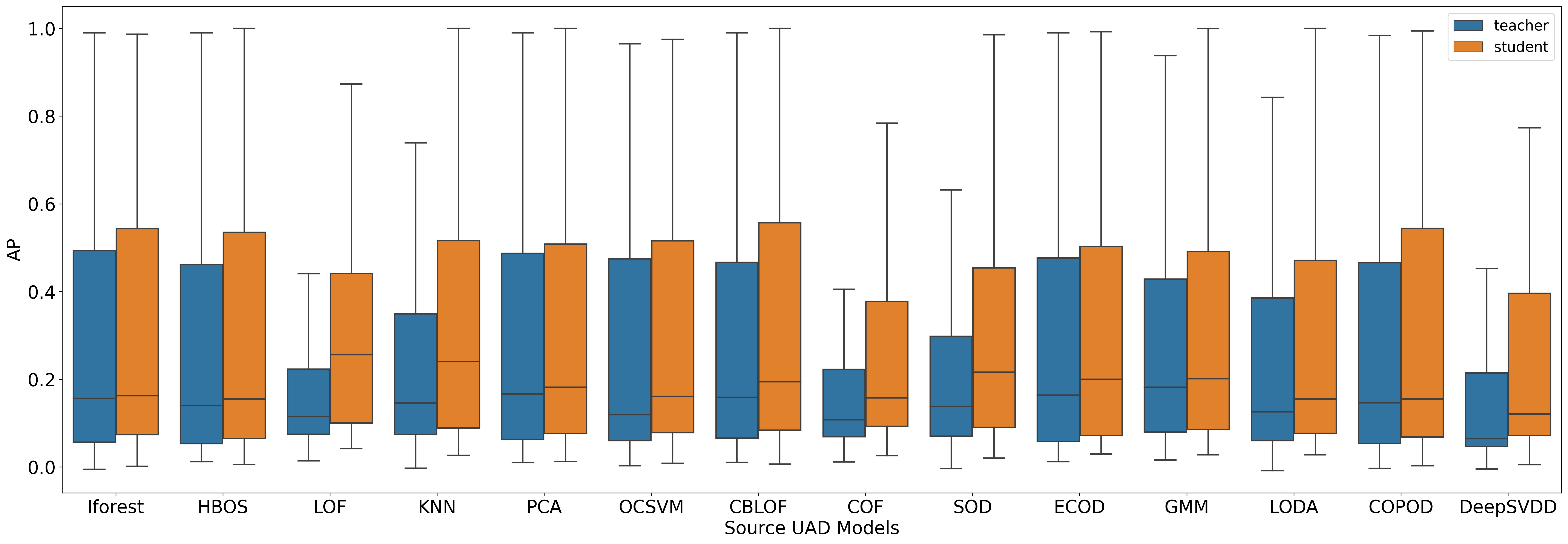}
\caption{Evaluation metric: Average Precision (AP)}
\label{fig:three sin x}
\end{subfigure}
\caption{
Boxplots of performance of 14 teacher UAD models and their UADB boosters on 84 tabular datasets.
}
\label{fig:box overall}
\vspace{-1em}
\end{figure}

\textbf{Ablation Study (RQ3).}
To further validate the effectiveness of the error correction mechanism in UADB, we carry out ablation study on real-world datasets.
Note that in our case, removing error correction in UADB results in no knowledge correction, which forces the booster models to have identical output with the teacher model.
So the ablation study can be done by simply comparing the performance of teacher UAD models and their UADB boosters.
Same as above, to obtain reliable conclusions, we conduct the ablation study on all 84 real-world tabular datasets.
We summarize the ablation study results in boxplots of performance, as shown in Fig.~\ref{fig:box overall}.

We can observe that: 
\begin{itemize}
    \item In terms of both evaluation metrics, removing the error correction mechanism results in obvious average performance degradation for all tested UAD techniques.
    \item The degradation is more significant in terms of AP, where not only the average, but also the best and worst performance among the 84 datasets get degraded.
    \item The performance gain from error correction is especially important for those UAD models that do not perform well by themselves, such as LOF~\cite{breunig2000lof}, COF~\cite{tang2002enhancing}, KNN~\cite{ramaswamy2000efficient}, SOD~\cite{kriegel2009outlier}, and DeepSVDD~\cite{ruff2018deep} (in terms of AP).
\end{itemize}

\begin{table*}[t!]
\footnotesize
\centering
\caption{Ablation study results. The average AUCROC and AP over 84 datasets are reported. For each UAD model, we show the results of 6 variants (UAD itself and 5 different types of booster). The best results are highlighted in bold.}
\label{tab:ablation study}
\setlength{\tabcolsep}{0.7mm}{
\begin{tabular}{c|cccccccccccccc|c} 
\toprule
\multicolumn{16}{c}{\textbf{Comparison between different booster training strategies in terms of AUCROC}} \\ 
\midrule
\textbf{Source UAD Model}                 & \textbf{Iforest}          & \textbf{HBOS}          & \textbf{LOF}          & \textbf{KNN}          & \textbf{PCA}          & \textbf{OCSVM}          & \textbf{CBLOF}          & \textbf{COF}          & \textbf{SOD}          & \textbf{ECOD}          & \textbf{\textbf{GMM}}                   & \textbf{LODA}          & \textbf{COPOD}          & \textbf{\textbf{DeepSVDD}} & \textbf{Average}  \\ 
\midrule
\textbf{Origin}                  & 0.7028                    & 0.6848                 & 0.6311                & 0.6794                & 0.6930                & 0.6750                  & 0.7110                  & 0.6105                & 0.6638                & 0.6866                 & 0.7274                                  & 0.6571                 & 0.6882                  & 0.5346                     & 0.6777            \\ 
\hline
\textbf{Na\"ive Booster}           & 0.6896                    & 0.6728                 & 0.6365                & 0.6778                & 0.6943                & 0.6733                  & 0.7093                  & 0.6108                & 0.6813                & 0.6863                 & 0.7173                                  & 0.6751                 & 0.6713                  & 0.5641                     & 0.6766            \\ 
\hline
\textbf{Discrepancy Booster}    & 0.5589                    & 0.5526                 & 0.5807                & 0.5850                & 0.6065                & 0.6019                  & 0.5356                  & 0.5439                & 0.6025                & 0.5497                 & 0.6362                                  & 0.5677                 & 0.5607                  & 0.5851                     & 0.5755            \\ 
\hline
\textbf{Self Booster}        & 0.6838                    & 0.6688                 & 0.6334                & 0.6686                & 0.6809                & 0.6618                  & 0.7008                  & 0.6073                & 0.6701                & 0.6722                 & 0.7031                                  & 0.6736                 & 0.6658                  & 0.5667                     & 0.6685            \\ 
\hline
\textbf{Discrepancy Booster*} & 0.6235                    & 0.5904                 & 0.5731                & 0.6144                & 0.6195                & 0.5970                  & 0.5679                  & 0.5670                & 0.6358                & 0.6122                 & 0.6597                                  & 0.5761                 & 0.6039                  & 0.5580                     & 0.6031            \\ 
\hline
\textbf{UADB}                    & \textbf{0.7144}           & \textbf{0.7001}        & \textbf{0.7004}       & \textbf{0.7245}       & \textbf{0.7060}       & \textbf{0.6991}         & \textbf{0.7324}         & \textbf{0.6686}       & \textbf{0.7199}       & \textbf{0.7010}        & \textbf{\textbf{0.7407}}                & \textbf{0.6864}        & \textbf{0.6998}         & \textbf{\textbf{0.6343}}   & \textbf{0.7072}   \\ 
\bottomrule
\multicolumn{1}{l}{}             & \multicolumn{1}{l}{}      & \multicolumn{1}{l}{}   & \multicolumn{1}{l}{}  & \multicolumn{1}{l}{}  & \multicolumn{1}{l}{}  & \multicolumn{1}{l}{}    & \multicolumn{1}{l}{}    & \multicolumn{1}{l}{}  & \multicolumn{1}{l}{}  & \multicolumn{1}{l}{}   &                                         & \multicolumn{1}{l}{}   & \multicolumn{1}{l}{}    & \multicolumn{1}{l}{}       &                   \\ 
\toprule
\multicolumn{16}{c}{\textbf{\textbf{Comparison between different booster training strategies in terms of AP}}}\\ 
\midrule
\textbf{Source UAD Model}                 & \textbf{\textbf{Iforest}} & \textbf{\textbf{HBOS}} & \textbf{\textbf{LOF}} & \textbf{\textbf{KNN}} & \textbf{\textbf{PCA}} & \textbf{\textbf{OCSVM}} & \textbf{\textbf{CBLOF}} & \textbf{\textbf{COF}} & \textbf{\textbf{SOD}} & \textbf{\textbf{ECOD}} & \textbf{\textbf{\textbf{\textbf{GMM}}}} & \textbf{\textbf{LODA}} & \textbf{\textbf{COPOD}} & \textbf{\textbf{DeepSVDD}} & \textbf{Average}  \\ 
\midrule
\textbf{Origin}                  & 0.3012                    & 0.2918                 & 0.1903                & 0.2550                & 0.3051                & 0.2738                  & 0.3057                  & 0.1989                & 0.2322                & 0.2908                 & 0.2805                                  & 0.2623                 & 0.2832                  & 0.1727                     & 0.2670            \\ 
\hline
\textbf{Na\"ive Booster}           & 0.2966                    & 0.2926                 & 0.2257                & 0.2572                & 0.2923                & 0.2735                  & 0.3086                  & 0.2154                & 0.2608                & 0.2838                 & 0.2901                                  & 0.2865                 & 0.2841                  & 0.2187                     & 0.2744            \\ 
\hline
\textbf{Discrepancy Booster}    & 0.1588                    & 0.1629                 & 0.1739                & 0.1837                & 0.1946                & 0.1854                  & 0.1580                  & 0.1647                & 0.1881                & 0.1537                 & 0.2080                                  & 0.1813                 & 0.1627                  & 0.1856                     & 0.1751            \\ 
\hline
\textbf{Self Booster}        & 0.2946                    & 0.2911                 & 0.2127                & 0.2594                & 0.2938                & 0.2760                  & 0.3095                  & 0.2132                & 0.2647                & 0.2837                 & 0.3033                                  & 0.2887                 & 0.2817                  & 0.2132                     & 0.2748            \\ 
\hline
\textbf{Discrepancy Booster*} & 0.2078                    & 0.2057                 & 0.1854                & 0.2076                & 0.2255                & 0.1999                  & 0.1825                  & 0.1802                & 0.2260                & 0.1940                 & 0.2564                                  & 0.1897                 & 0.2015                  & 0.1988                     & 0.2048            \\ 
\hline
\textbf{UADB}                    & \textbf{0.3146}           & \textbf{0.3055}        & \textbf{0.3049}       & \textbf{0.3178}       & \textbf{0.3061}       & \textbf{0.2967}         & \textbf{0.3241}         & \textbf{0.2659}       & \textbf{0.3064}       & \textbf{0.3009}        & \textbf{\textbf{0.3092}}                & \textbf{0.3013}        & \textbf{0.2978}         & \textbf{\textbf{0.2468}}   & \textbf{0.3039}   \\
\bottomrule
\end{tabular}}
\end{table*}

\textbf{Comparison with Other Intuitive Mechanisms (RQ4).}
Finally, to further validate the effectiveness of UADB's design, we consider several variants of UADB and compare their performance.
Motivated by previous works that directly use the discrepancy between multiple models' output as the predicted anomaly score, we apply different training and inference schema to generate multiple variant booster frameworks.
Specifically, we consider 4 different alternative booster frameworks:
\begin{enumerate}
    \item Na\"ive Booster: it only uses the source model's output as static pseudo labels, then use it to train the booster model without any in-iteration adjustments on pseudo labels. The booster's output are used directly as the predicted anomaly score at inference.
    \item Discrepancy Booster: similar to the Na\"ive Booster, it also adopt the source model's output as static labels for pseudo-supervised training of the booster model. However, when inference, it uses the discrepancy (Standard deviation) between booster's output and source model's output as the predicted anomaly score.
    \item Self Booster: it involves multiple booster training iterations like UADB, but in each iteration, it does not perform error correction on the pseudo labels. Instead, it uses booster's output after MinMax normalization as the pseudo label in the next iterations. At inference, booster's output is used as the predicted anomaly score.
    \item Discrepancy Booster*: it also involves multiple booster training iterations like UADB. The booster training strategy is identical to that of Self Booster, but at inference, the discrepancy between booster's output and source model's output is used as the predicted anomaly score.
\end{enumerate}

Following the previous setting, we apply all booster frameworks to 14 UAD models and 84 tabular datasets to get reliable results.
The comparison results between UADB and these alternative booster frameworks are shown in Table \ref{tab:ablation study}.
We can obeserve that:
\begin{itemize}
    \item UADB is the best performer among all booster frameworks, it generally outperforms other counterparts by a large margin.
    \item Directly use the discrepancy between a teacher and a student model's output as the anomaly score cannot get good anomaly detection performance. This may due to the fact that only two models are included causing inaccurate discrepancy estimation.
    \item Self Booster obtains better results than other counter parts, indicating the importance of multi-stage training and pseudo labels adjustments.
\end{itemize}

To this point, we answered all the research questions that were proposed at the start of this section.
Through comprehensive experiments and analysis on multiple real-world datasets and UAD models, we show that UADB can generally boost the performance of different UAD models on heterogeneous tabular datasets, and the error correction mechanism can effectively prevent the transfer of the wrong knowledge in the teacher model. UADB generally shows wide applicability and great performance in various real-world applications.

\section{Conclusion}
\label{sec:conclusion}
In this paper, we aim to design better unsupervised anomaly detection (UAD) techniques for tabular datasets.
This task faces several fundamental challenges including (i) unsupervision: lack of prior knowledge about the anomalous pattern; (ii) assumption misalignment: the assumptions of UAD methods can be easily violated in real-world data and result in suboptimal performance; (iii) data heterogeneity: tabular data's properties vary greatly across different domains.
For above reasons, a universal winner that consistently outperforms other solutions does not exist due to the multifaceted complexity of the task.
We argue that the key to generally better UAD on diverse and heterogeneous tabular data is to go beyond the static assumptions and empower UAD models with adaptability to different data.

In light of this, we propose \method (\underline{U}nsupervised \underline{A}nomaly \underline{D}etection \underline{B}ooster), a versatile framework for improving any UAD model's general performance on tabular datasets.
Specifically, UADB is designed to (i) keep the prior knowledge of the UAD model and its assumption by knowledge transfer, and (ii) perform adaptive error correction during transfer by exploiting the sample variance at the same time.
The structure of UADB is conceptually simple, but has proven to be very effective in real-world UAD tasks.
Extensive experiments show that UADB can generally achieve significant performance improvement over the 14 different source UAD models on 84 heterogeneous tabular datasets.
To our best knowledge, UADB is the first of its kind as a general framework for enhancing UAD models.
To sum up, we carry out a preliminary exploration on designing a model-agnostic booster framework to enhance UAD on tabular datasets.
We hope our findings can shed some light on developing better versatile UAD solutions.

\bibliographystyle{IEEEtran}
\bibliography{reference}
\vspace{12pt}
\end{document}